%% file: neurips_2025.tex
\documentclass{article}

\usepackage[preprint]{neurips_2025}

\usepackage[utf8]{inputenc} 
\usepackage[T1]{fontenc}    
\usepackage{hyperref}       
\usepackage{url}            
\usepackage{booktabs}       
\usepackage{amsfonts}       
\usepackage{nicefrac}       
\usepackage{microtype}      
\usepackage{xcolor}         
\usepackage{comment}        
\usepackage{amsthm}
\newtheorem{proposition}{Proposition}
\usepackage{amsmath}
\usepackage{amssymb}
\newtheorem{lemma}{Lemma}
\newtheorem{theorem}{Theorem}
\newtheorem{definition}{Definition}
\usepackage{enumitem}
\usepackage[most]{tcolorbox}
\usepackage{algorithm}
\usepackage{algpseudocode}
\usepackage{caption}
\usepackage{cleveref}
\usepackage{tikz}
\usepackage{tabularx}

\usepackage{graphicx}
\usepackage{wrapfig}
\usepackage{subcaption}

\title{Active Target Discovery under Uninformative Prior: The Power of Permanent and Transient Memory}

\author{
    Anindya Sarkar\thanks{Equal contribution} ,~~~~~~Binglin Ji\footnotemark[1] ,~~~~~~Yevgeniy Vorobeychik \\
    \texttt{\{anindya,~binglin.j,~yvorobeychik\}@wustl.edu,}\\Department of Computer Science and Engineering\\ Washington University in St. Louis, USA\\}

\begin{document}

\maketitle

\begin{abstract}
  In many scientific and engineering fields, where acquiring high-quality data is expensive—such as medical imaging, environmental monitoring, and remote sensing—strategic sampling of unobserved regions based on prior observations is crucial for maximizing discovery rates within a constrained budget. The rise of powerful generative models, such as diffusion models, has enabled active target discovery in partially observable environments by leveraging learned priors—probabilistic representations that capture underlying structure from data. With guidance from sequentially gathered task-specific observations, these models can progressively refine exploration and efficiently direct queries toward promising regions. However, in domains where learning a strong prior is infeasible due to extremely limited data or high sampling cost (such as rare species discovery, diagnostics for emerging diseases, etc.), these methods struggle to generalize. To overcome this limitation, we propose a novel approach that enables effective active target discovery even in settings with uninformative priors, ensuring robust exploration and adaptability in complex real-world scenarios. Our framework is theoretically principled and draws inspiration from neuroscience to guide its design. Unlike black-box policies, our approach is inherently interpretable, providing clear insights into decision-making. Furthermore, it guarantees a strong, monotonic improvement in prior estimates with each new observation, leading to increasingly accurate sampling and reinforcing both reliability and adaptability in dynamic settings. Through comprehensive experiments and ablation studies across various domains, including species distribution modeling and remote sensing, we demonstrate that our method substantially outperforms baseline approaches.
\end{abstract}

\input{intro}

\input{prob}

\input{method}

\input{exp}

\input{abl}

\input{rel_work}

\input{Conclusion}
\paragraph{Acknowledgement} This research was partially supported by the National Science Foundation (CCF-2403758, IIS-2214141), Army Research Office (W911NF2510059), Office of Naval Research (N000142412663), and Amazon.

\bibliographystyle{unsrt}
\bibliography{reference}

\newpage

\newpage
\appendix
\input{appendix}

\end{document}

%% file: intro.tex
\section{Introduction}\label{sec:int}
\vspace{-2pt}
Active Target Discovery (ATD) is a fundamental problem in scientific and engineering domains where data acquisition is expensive and environments are only partially observable. 
Beginning with an unobservable task, the goal is to strategically and sequentially sample glimpses to infer and uncover the underlying target, such as disease regions in medical imaging, pollution hotspots in environmental monitoring, or objects of interest in satellite imagery—all while adhering to a strict sampling budget.
This task is critical in scenarios like detecting rare tumors in MRI scans, localizing contaminants in remote areas, or identifying missing persons in search-and-rescue missions. In all these cases, the cost of each observation is high, and acquiring feedback from the ground-truth reward function often requires expert judgment, specialized equipment, and substantial time.
Diffusion-guided Active Target Discovery (DiffATD)~\citep{sarkar2025onlinefeedbackefficientactive} has been recently proposed as a promising solution to the ATD task. It operates by learning a strong prior over the search space using samples from the domain of interest. Specifically, DiffATD~\citep{sarkar2025onlinefeedbackefficientactive} leverages a diffusion model trained on domain-specific samples to serve as a strong prior, guiding its sampling strategy via learned diffusion dynamics conditioned on the gathered observations to efficiently uncover target regions within a constrained sampling budget. This prior-driven approach allows it to balance exploration and exploitation in a principled way, often outperforming baseline methods in well-studied domains. However, the effectiveness of DiffATD~\citep{sarkar2025onlinefeedbackefficientactive} fundamentally hinges on the availability of representative prior samples from the target domain. In many real-world scenarios—such as emerging diseases, rare environmental hazards, or underexplored geographic regions—such samples are either unavailable or prohibitively difficult to obtain. In these cases, the strong prior assumption breaks down, and methods like DiffATD~\citep{sarkar2025onlinefeedbackefficientactive} struggle to generalize, limiting their applicability in precisely the settings where efficient target discovery is most critical.

To confront the challenge of ATD without access to prior samples, we draw inspiration from Neuroscience. The real world is inherently dynamic, continually presenting novel and unforeseen situations. Yet, humans navigate such environments seamlessly by drawing on two complementary systems: a long-term memory that encodes structured, generalizable knowledge, and a short-term memory that rapidly adapts to immediate context. For instance, when driving in an unfamiliar city, we rely on our understanding of traffic rules (long-term knowledge) while dynamically responding to unexpected road closures or detours (short-term adaptation). 
Kumaran~\cite{kumaran2012representations} suggests that this dual mechanism is rooted in the neocortex and hippocampus, that is, the brain leverages a dual memory system comprised of \emph{permanent memory}—mediated by the neocortex—which encodes structured knowledge through slow learning and supports generalization, and \emph{transient memory}—associated with the hippocampus—which enables rapid acquisition of precise, task-specific information. 
Inspired by this cognitive mechanism, we argue that emulating such a dual memory structure in artificial agents is essential for solving new tasks in dynamic environments, where both stable generalization and fast adaptability are crucial.
\begin{figure}
    \centering
    \includegraphics[height=4.4cm,width=0.98\textwidth]{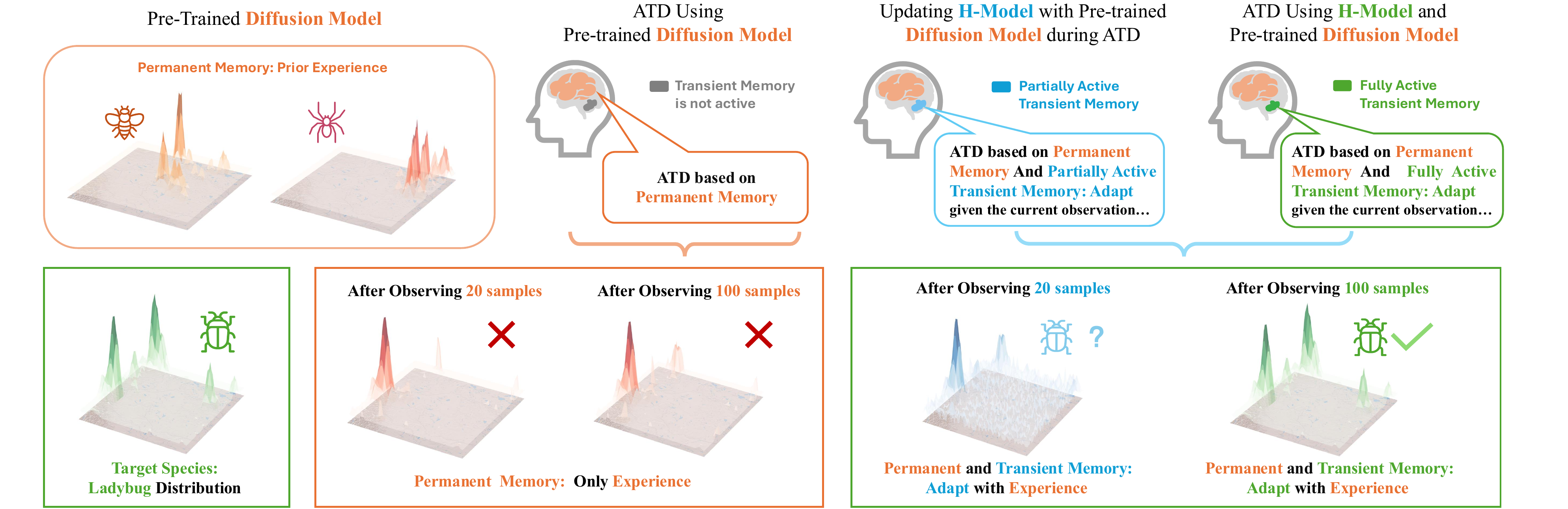}
    \caption{Interplay of Permanent and Transient Memory to Guide Active Target Discovery.}
    \label{fig:intro}
    \vspace{-6pt}
\end{figure}

Inspired by this mechanism, we draw on Bayesian inference to tackle ATD in the absence of prior domain data. At the heart of our approach lies an Expectation-Maximization-style algorithm that guarantees monotonic improvement of the sampling prior—modeled as a diffusion process—as new observations unfold across the environment. This evolving prior actively guides the sampling strategy to maximize discovery under a strict budget. Our prior model architecture reflects mechanisms from the brain’s dual-memory system by pairing a powerful \emph{pretrained diffusion model} as a long-term or \emph{permanent memory}—capturing rich, generalizable structure across domains—with a lightweight, adaptive module based on \emph{Doob’s h-transform} that serves as \emph{transient memory}, enabling rapid contextual adaptation from new observations. 
The lightweight Doob’s h-transform module is updated after every few observations, allowing it to swiftly adapt to the evolving dynamics of the current task. This synergy enables our system to combine the power of global knowledge with the flexibility of fast local adaptation, allowing for efficient, intelligent exploration in novel and data-scarce environments. Crucially, the integration of a pretrained diffusion model with a lightweight network that implements \emph{Doob’s h-transform} enables principled sampling from the posterior distribution conditioned on the accumulated observations. This design ensures that each sampling decision is grounded in both prior knowledge and task-specific context, lending theoretical rigor and practical effectiveness to our architecture. Finally, our sampling strategy ranks unobserved points by balancing exploration and exploitation scores, driven by an updated prior and an online-trained reward model that learns the target's characteristics from gathered observations. As more observations are revealed, the prior progressively improves, leading to more accurate rankings that fuel a highly informed active discovery process, reinforced by theoretical guarantees and compelling empirical evidence. Figure~\ref{fig:intro} illustrates the core motivation behind our approach. We summarize our contributions as follows:
\vspace{-7pt}
\begin{itemize}[noitemsep,topsep=0pt, leftmargin=*]
    \begin{tcolorbox}[colback=blue!5!white, colframe=blue!40!black, boxrule=0.5pt, arc=2mm]
    \item We introduce a novel and principled framework, Expectation Maximized Permanent Temporary Diffusion Memory (EM-PTDM), designed to uncover targets of interest within a strict sampling budget in partially observable environments. Unlike prior methods, EM-PTDM operates without relying on task-specific prior domain data, significantly broadening its applicability. Furthermore, it integrates a white-box, interpretable sampling policy grounded in Bayesian experiment design, enabling transparent and strategically guided exploration.
    \item EM-PTDM framework is backed by rigorous empirical evidence and ensures a monotonic refinement of the prior as new observations are gathered. This continual refinement directly enhances the accuracy of the scoring mechanism used for sampling, resulting in more precise and efficient active target discovery over time.
    \item We demonstrate the significance of each component in our proposed EM-PTDM method through extensive quantitative and qualitative ablation studies across a range of datasets, including species distribution modeling and remote sensing.
    \end{tcolorbox}
\end{itemize}
\vspace{-3pt}

%% file: prob.tex
\section{Problem Formulation}
\vspace{-6pt}
In this section, we present the details of our proposed Active Target Discovery (ATD) task setup. 
  ATD involves actively uncovering one or more targets within a search area, represented as an (initially unobserved) region $x$ divided into $N$ grid cells, such that $x = (x^{(1)}, x^{(2)}, ..., x^{(N)})$. ATD operates under a query budget $\mathcal{B}$, representing the maximum number of measurements allowed. Each grid cell represents a sub-region and serves as a potential measurement location. A measurement of location $i$ \emph{provides feedback}, revealing both \emph{the content} of a specific sub-region $x^{(i)}$ for the $i$-th grid cell, as well as yielding 
  \emph{an outcome} $y^{(i)}$ $\in$ $[0, 1]$, where $y^{(i)}$ is the fraction of pixels in the grid cell $x^{(i)}$ that belong to the target of interest.
In each task configuration, the target's content is initially unknown and is revealed incrementally through observations from measurements. The goal is to identify as many grid cells belonging to the target as possible by strategically exploring the grid within the given budget $\mathcal{B}$.
Denoting a query performed in step $t$ as $q_t$, the overall task optimization objective is:
\begin{equation}\label{eq:objective}
    U(x^{\{(q_t)\}};\{q_t\}) \equiv 
    \max_{\{q_t\}} \sum_{t} y^{(q_t)} \quad\\
    \text{ subject to } \> t \le \mathcal{B}
\end{equation}
With objective~\ref{eq:objective} in focus, we aim to develop a search policy that efficiently explores the search area ($x_\text{test} \sim X_\text{test}$) to identify as many target regions as possible within a measurement budget $\mathcal{B}$— \textbf{all without requiring access to any prior samples from the domain of interest.} 
\vspace{-7pt}


%% file: method.tex
\section{Methodology}\label{sec:method}
\vspace{-6pt}
A central challenge in ATD under an initially uninformative prior lies in achieving three distinct goals: (1) progressively improving the prior as sequential observations accumulate—since the prior fundamentally shapes the ATD process; (2) designing the prior to be swiftly adaptable, capable of rapidly incorporating knowledge from limited observations, as demanded by the task's high data-efficiency requirements, and (3) utilizing the sequentially improved prior model to strategically sample unobserved regions to maximize target discovery within limited sampling budget. 
\vspace{-5pt}
\paragraph{Progressively Improving the Prior model} We begin by addressing the crucial challenge of enabling the prior to evolve progressively—ensuring that each newly acquired observation from the search space contributes meaningfully to refining the model’s understanding and guiding sampling more effectively. To this end, we formulate our approach within the principled structure of the Expectation-Maximization (EM) framework, utilizing its iterative inference and optimization steps to systematically refine the prior with each new observation. Specifically, our goal is to learn the parameters of a prior model (e.g., a DDPM), denoted as $q_{\phi}(x, y)$, by maximizing the log-evidence $\log q_{\phi}(y)$ for a given observation $y$. Considering a distribution over observations $p(y)$, this corresponds to maximizing the expected log-evidence, which is equivalent to minimizing the KL divergence between the true distribution $p(y)$ and $q_{\phi}(y)$:
\begin{equation}
\phi^* = \arg\max_{\phi} \mathbb{E}_{p(y)}[\log q_{\phi}(y)]= \arg\min_{\phi} \mathrm{KL}(p(y) \| q_{\phi}(y)).
\end{equation}

Next, we derive an iterative optimization procedure for $\phi$ that ensures monotonic improvement of expected log-evidence $\mathbb{E}_{p(y)}[\log q_{\phi}(y)]$. We present the result in the following Proposition. 

\begin{tcolorbox}[colback=blue!5!white, colframe=blue!40!black, boxrule=0.2pt, arc=1mm, width=\textwidth]
\begin{proposition}\label{pr:1}
Let $\phi_k$ denote the parameters of the current prior model, then improving this prior by maximizing the expected log-evidence 
$\mathbb{E}_{p(y)}[\log q_{\phi_k}(y)]$ with respect to $\phi_k$ is equivalent to maximizing the following surrogate maximization:
\vspace{-2pt}
\begin{equation}\label{eq:em}
\phi_{k+1} = \arg\max_{\phi} \, \mathbb{E}_{p(y)} \, \mathbb{E}_{\text{\colorbox{blue!10}{$q_{\phi_k}(x|y)$}}} \left[ \log q_{\phi}(x) \right]
\end{equation}
such that $\mathbb{E}_{p(y)}[\log q_{\phi_{k+1}}(y)] \geq \mathbb{E}_{p(y)}[\log q_{\phi_{k}}(y)]$. Where $q_{\phi_k}(x|y)$ represents the approximate posterior under the current prior $\phi_k$. Thus, this update ensures a strict improvement in the expected log-evidence. We present detailed proof in the Appendix.
\end{proposition}
\end{tcolorbox}
\vspace{-3pt}
As established in~\ref{pr:1}, optimizing the prior using the objective in Eqn.~\ref{eq:em} ensures that each update step systematically refines the prior toward a more faithful approximation of the underlying data distribution. Specifically, $q_{\phi_{k+1}}(x)$ is more consistent with the distribution of observations $p(y)$ than $q_{\phi_{k}}(x)$. The update of the prior parameters $\phi_{k+1}$ follows a two-stage procedure: first, samples are drawn from the posterior distribution $q_{\phi_k}(x|y)$; next, the prior model $q_{\phi_{k+1}}(x)$ is trained to approximate this posterior by fitting to the generated samples. A central challenge in this framework is the accurate sampling from the posterior distribution $q_{\phi_k}(x|y)$, particularly during the early stages of active discovery when observations are inherently sparse. This data scarcity significantly limits the efficacy of traditional data-driven optimization approaches. Moreover, the challenge is intensified by the fact that repeated updates ($k > 1$) are both computationally and time-demanding. Consequently, it is imperative to design a prior model that is both swiftly adaptable to limited observation and lightweight in complexity. Large-scale diffusion models, while powerful, are typically data-intensive and ill-suited for such low-resource settings. Our objective is \emph{to develop a model that can efficiently adapt under constrained data regimes without compromising the quality of posterior approximation.}
\vspace{-9pt}
\paragraph{Swiftly Adaptable Prior model with Limited Observation} To address this challenge, we 
draw inspiration from Doob's $h$-transform, which enables sampling conditioned on the observations gathered. This perspective offers a powerful lens through which we construct an adaptive mechanism capable of efficiently guiding the prior with minimal data. Specifically, conditional sampling from the posterior distribution $q_{\phi_k}(x|y)$ requires access to the posterior score $\nabla_{x} \log p_t(x \mid Y = y)$, conditioned on a given observation $y$. Since diffusion models are trained to approximate the score function of the marginal data distribution, i.e., $s^{\theta^*}_t \approx \nabla_x \log p_t(x)$, a pre-trained diffusion model can be adapted for conditional inference by applying Bayes' theorem:
\vspace{-4pt}
\begin{equation}\label{eq:bayes}
    \nabla_x \log p_t(x \mid Y = y) \approx s^{\theta^*}_t + \nabla_x \log p_t(Y = y \mid x).
\end{equation}
The term $\nabla_x \log p_t(Y = y \mid x)$ is commonly referred to as the \emph{guidance} term, as it effectively steers the reverse diffusion process toward samples consistent with the observation $y$. However, computing this guidance term is generally analytically intractable, which poses a significant challenge in performing conditional generation.
Before delving into how Doob's $h$-transform can be harnessed to efficiently approximate the conditional guidance term, especially in data-scarce regimes, we first introduce the formal definition of Doob's $h$-transform. This foundational understanding will later enable us to leverage its powerful structure for efficient and adaptive conditional sampling.

\begin{tcolorbox}[colback=blue!5!white, colframe=blue!40!black, boxrule=0.2pt, arc=1mm]
\begin{definition}[Doob's \textit{h}-transform]~\citep{rogers2000diffusions2,heng2021simulating}
\label{prop:doob}
Consider the following unconditional reverse-time SDE with $f_t$ as the drift and $\sigma_t$ as the diffusion coefficient.
\[
\small{dX_t = \left( f_t(X_t) - \sigma_t^2 \nabla_{X_t} \ln p_t(X_t) \right) dt + \sigma_t dW_t, \quad X_T \sim P_T \sim \mathcal{N}(0,1)}
\]
The corresponding conditional process $(X_t \mid X_0 \in B)$ evolves according to the SDE:
\begin{equation}\label{eq:doob}
    \small{dH_t = \left[  \overleftarrow{b_t(H_t)} - \sigma_t^2 
    \, \colorbox{blue!10}{$\displaystyle \nabla_{H_t} \ln p_{0|t}(X_0 \in B \mid H_t)$} \right] dt 
    + \sigma_t \overleftarrow{dW_t}, \quad H_T \sim \mathcal{P}_T \sim \mathcal{N}(0,1)}
\end{equation}
where the backward drift $\overleftarrow{b_t(H_t)}$ is defined as: $ \overleftarrow{b_t(H_t)} = f_t(H_t) - \sigma_t^2 \nabla_{H_t} \ln p_t(H_t)$,
and the conditional law satisfies $\mathrm{Law}(H_s \mid H_t) = \overrightarrow{p}_{s|t,0}(x_s \mid x_t, x_0 \in B)$ with $\mathbb{P}(X_0 \in B) = 1$.

We denote the conditional process as $H_t$ and the corresponding unconditional process as $X_t$. Doob’s \textit{h}-transform illustrates that conditioning a diffusion process to reach a target set $X_0 \in B$ at terminal time $T$ yields a new diffusion process governed by an adjusted drift term (as shown in \colorbox{blue!10}{blue}). This conditional process is guaranteed to reach the desired event within a finite time $T$. The function $h(t, H_t) \triangleq \overleftarrow{{p}_{0|t}}(X_0 \in B \mid H_t)$ is known as the \textit{h}-transform.
\end{definition}
\end{tcolorbox}

Definition~\ref{prop:doob} highlights a key insight: conditional sampling, informed by the observations collected up to the current step, hinges on the interplay between the unconditional score and a conditional score term, which is solely characterized by Doob’s h-transform. Consequently, comparing Equation~\ref{eq:bayes} and~\ref{eq:doob}, the posterior score can be formulated as:
\vspace{-2pt}
\begin{equation}\label{eq:memory}
    \small{\nabla_x \log p_t(x \mid Y = y) \approx \underbrace{s^{\theta^*}_t(x)}_{\text{Permanent Memory}} + \underbrace{\nabla_x \log p_t(Y = y \mid x)}_{\text{$\approx h_t^{\zeta}(x,y)$: Transient Memory}}.}
\end{equation}
\vspace{-2pt}

As shown in Equation~\ref{eq:memory}, the posterior score naturally decomposes into two complementary components. The first term, derived from a pretrained diffusion model, serves as a form of \emph{permanent memory}—encapsulating broad, generalizable patterns acquired from large-scale training data. While this offers a strong prior, it alone is insufficient in our setting, where no access to domain-specific prior samples is available. To address this limitation, we introduce the second term as a \emph{transient memory} that modulates the pretrained dynamics in response to the limited observations collected during the active discovery process. This adaptive correction is governed by Doob’s $h$-transform, parameterized by $\zeta$, allowing the model to align with newly encountered observations in the data. In the Appendix, we provide an interpretation that frames the $h$-transform as a correction mechanism for the unconditional score, offering insights into how it adapts the pretrained dynamics to the conditional setting. Next, we detail how $\zeta$ can be efficiently learned to enable rapid adaptation in data-scarce scenarios. Interestingly, the parameter $\zeta$ can be optimized using an approach analogous to denoising score matching, as formalized in the following Lemma.
\vspace{-2pt}
\begin{tcolorbox}[colback=blue!5!white, colframe=blue!40!black, boxrule=0.2pt, arc=1mm]
\begin{lemma}
\label{eq:lem}
Consider the following stochastic differential equation for the conditional process:
\vspace{-2pt}
\[
    dH_t = \left[ f_t(H_t) - \sigma_t^2 \left( \nabla_{H_t} \ln p_t(H_t) + h_t(H_t) \right) \right] dt + \sigma_t dW_t,
\]
where $H_T \sim \mathcal{Q}^{f_t}_T[p(x_0 \mid y)] = \int p_{T|0}(x \mid x_0)\, p(x_0 \mid y)\, dx_0
$. The $h$-transform function $h^*_t$ admits a denoising score-matching representation:
 $h^*_t = \arg\min_{h_t \in \mathcal{H}} \mathcal{L}_{\mathrm{DSM}}(h_t)$, where 
    \[
        \mathcal{L}_{\mathrm{DSM}}(h_t) := \mathbb{E}_{\substack{X_0 \sim p(x_0 \mid y) \\ t \sim \mathcal{U}(0, T); H_t \sim p_{t|0}(x_t \mid x_0)}} \small{\left[ 
        \left\| \left( h_t(H_t) + \nabla_{H_t} \ln p_t(H_t) \right) - \nabla_{H_t} \ln p_{t|0}(H_t \mid X_0) \right\|^2 \right].}
    \]
\end{lemma}
\end{tcolorbox}
The proof is in the Appendix. Based on this Lemma, we optimize $\zeta$ as follows:
\begin{equation}\label{eq:sm}
\small{\min_{\zeta} \, \mathbb{E}_{(X_0, Y),\, \varepsilon,\, t} \left\| \left( h^{\zeta}_t(H_t, Y) + s^{\theta^*}_t(H_t) \right) - \varepsilon \right\|^2}.
\end{equation}
\noindent
Let $H_t = \sqrt{\bar{\alpha}_t} \, X_0 + \sqrt{1 - \bar{\alpha}_t} \, \varepsilon$, where $(X_0, Y) \sim q_{\phi}(x_0, y)$ and $\varepsilon \sim \mathcal{N}(0, \mathbf{I})$. The function $h^\zeta_t$ denotes a neural network designed to approximate the $h$-transform. Crucially, the loss function in Equation~\ref{eq:sm} operates exclusively through forward evaluations of the pre-trained model and does not necessitate backpropagation through the fixed parameters $\theta^\ast$. Consequently, the $h$-transform is tasked solely with learning the residual noise component, effectively serving as a corrective mechanism. This design choice permits the $h$-model to remain lightweight and shallow, thereby enabling rapid adaptation even in regimes with extremely limited observations. Following each observation, we draw posterior samples $X_0 \sim q_{\phi}(x_0 \mid y)$ using the posterior score composed of the frozen primary memory $s^{\theta^*}_t(x)$ and the transient memory $h^{\zeta}_t(x, y)$. These samples are then used to update the transient memory parameters $\zeta$ via the objective in~\ref{eq:sm}, thereby ensuring that the resulting prior $\phi$—consisting of both $\theta^*$ and the updated $\zeta$—yields a strict improvement in the expected log-evidence, as formalized in~\ref{eq:em}.
Formally, the following theorem establishes that updating the transient memory parameter $\zeta$ according to~\ref{eq:sm} leads to a monotonic improvement of the composite prior $\phi = (\zeta, \theta)$.
\vspace{-7pt}
\begin{tcolorbox}[colback=blue!5!white, colframe=blue!40!black, boxrule=0.2pt, arc=1mm]
\begin{theorem}
Let $\phi = (\theta, \zeta)$ denote the parameters of the prior, where $\theta$ corresponds to a fixed pre-trained model and $\zeta$ parameterizes the learnable $h$-transform module. Suppose $\zeta$ is updated according to Equation~\ref{eq:sm}, yielding an updated parameter $\zeta_{\text{new}}$ and an updated prior $\phi_{\text{new}} = (\theta, \zeta^{\text{new}})$. Then, the marginal expected log-evidence satisfies the following:
\[
   \mathbb{E}_{p(y)}[\log q_{\phi_{\text{new}}}(y)] \geq \mathbb{E}_{p(y)}[\log q_{\phi}(y)].
\]
\end{theorem}
\end{tcolorbox}
\vspace{-3pt}
We present detailed proof in the Appendix. Next, we present our methodology for updating the h-model, designed to enable a seamless adaptation of the reverse diffusion dynamics.
\vspace{-12pt}
\paragraph{Learning Dynamics of $h$-model}
A natural strategy is to iteratively update the h-model after each new observation, thereby refining the prior and leveraging this improved prior to guide subsequent rounds of ATD more effectively.  
However, in the early stages of the discovery process, the available observations are often too sparse to capture the structure of the underlying search space. Updating the h-model solely based on such limited data can lead to premature convergence to suboptimal regions in the parameter space, from which recovery becomes increasingly difficult—even as more data becomes available gradually as search progresses. This challenge is further intensified by
\begin{wrapfigure}{r}{0.4\textwidth}
  \centering
  \includegraphics[width=0.38\textwidth]{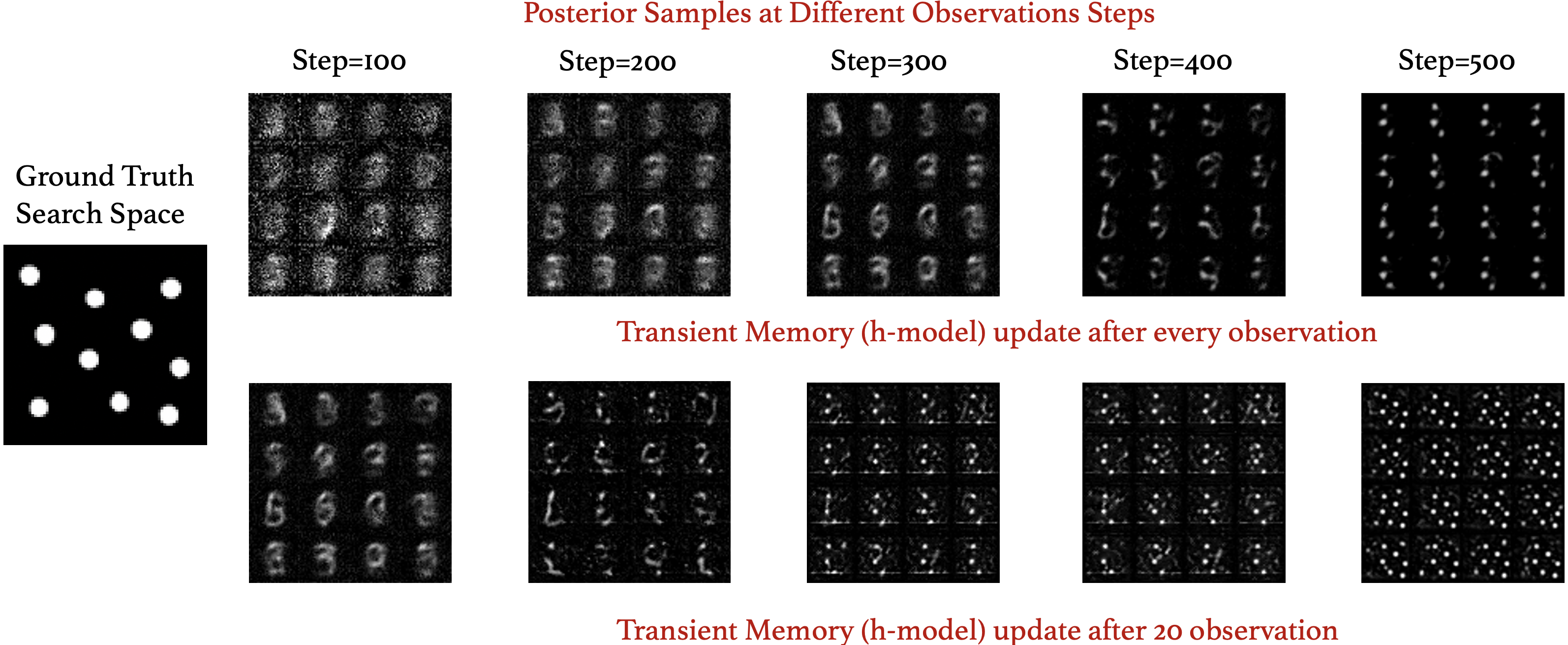}
  \caption{\small{$h$-model adaptability dynamics. We use a Diffusion model trained on MNIST as permanent memory (16 samples per Fig).}
  \label{fig:h}
}
  \label{fig:wrapped}
\end{wrapfigure}
 the $h$-model’s intended adaptability, as rapid updates may amplify the risk of overfitting to early, uninformative signals. To validate this hypothesis, we conducted a toy experiment (illustrated in Fig.~\ref{fig:h}), comparing two strategies for updating the h-model. In the first, the h-model is updated after every observation; in the second, updates occur only after accumulating several observations. We visualize the resulting posterior samples at different stages. When the h-model is updated after each individual observation, the posterior becomes noticeably noisy during the early phase of discovery, likely due to limited data. In contrast, postponing the updates results in more stable and coherent posterior samples that better reflect the gathered observations, enabling the diffusion process to incorporate new information more effectively at each stage.
Motivated by this observation, we introduce a simple yet effective scheduling strategy for updating the h-model. The proposed scheduler (detailed in Appendix) adapts dynamically with the query budget—starting with infrequent updates and increasing in frequency as the search progresses. This design leverages the fact that updates become more reliable over time due to the increasing accumulation of informative observations. 

\begin{figure}[!h]
\includegraphics[width=\linewidth]{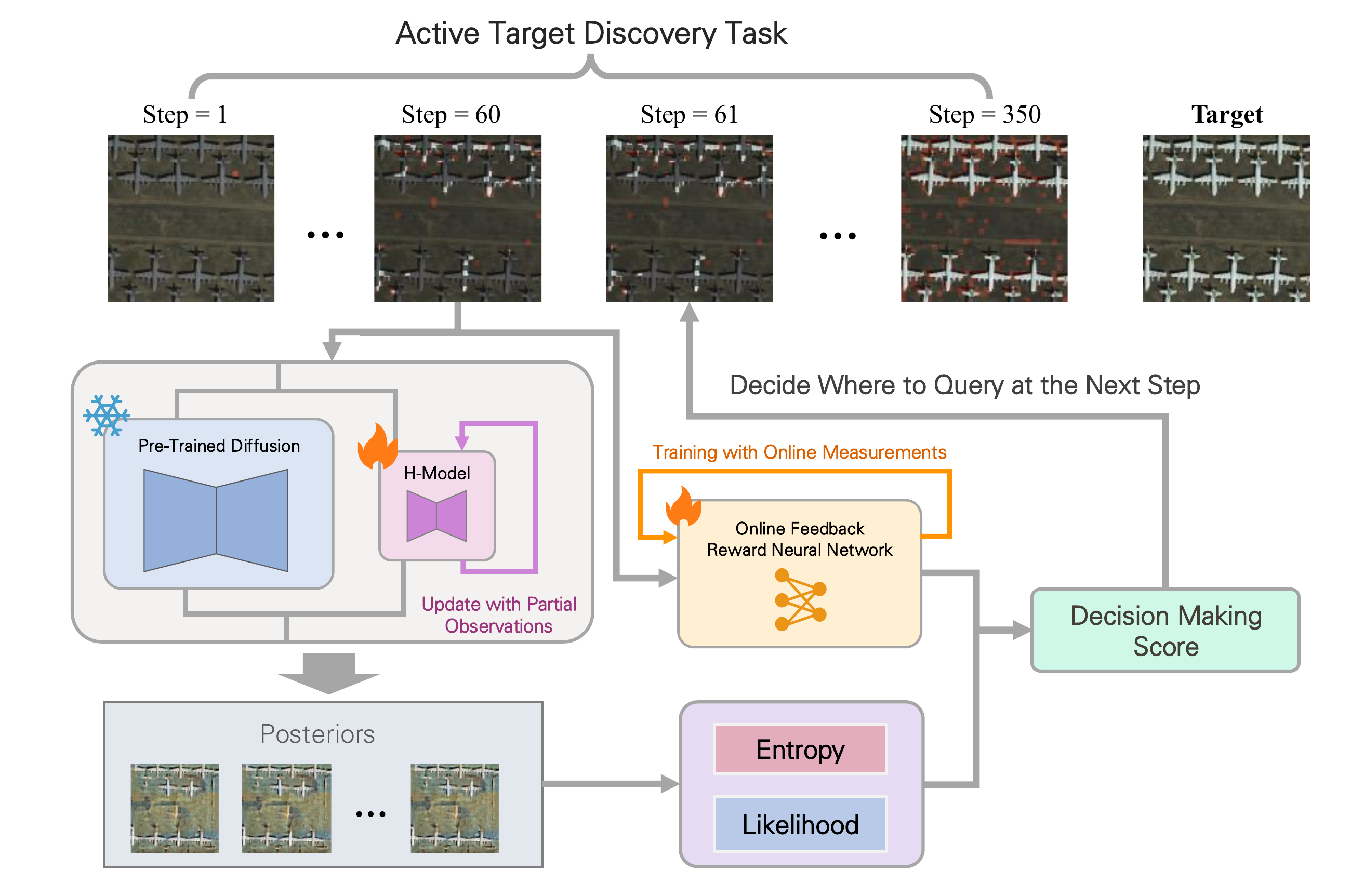}
\caption{An Overview of EM-PTDM Framework.}
\label{fig:PTDM_fm1}
\end{figure}

\vspace{-5pt}
\begin{tcolorbox}[colback=blue!5!white, colframe=blue!40!black, boxrule=0.2pt, arc=1mm]
\begin{theorem}
Let $\phi_t = (\theta, \zeta_t)$ be the prior parameters at observation step $t$. 
Suppose $\zeta_t$ is updated following Equation~\ref{eq:sm}
, yielding $\phi_{t+k} = (\theta, \zeta_{t+k})$ after $k$ additional observations. Then the expected log-evidence improves monotonically:
\[
\mathbb{E}_{p(y_{t+k})}[\log q_{\phi_{t+k}}(y)] \geq \mathbb{E}_{p(y_{t+k})}[\log q_{\phi_t}(y)].
\]
where $y_{t+k}$ represents the set of observations gathered uptill time $t+k$, and $k \geq 1$.
\end{theorem}
\end{tcolorbox}
\vspace{-4pt}
We present detailed proof in the Appendix. Next, we introduce a sampling strategy that capitalizes on the updated prior developed thus far, guiding effective and budget-conscious ATD.
\vspace{-9pt}
\paragraph{Sampling Strategy for ATD Utilizing the Updated Prior}
An effective sampling strategy that strikes the right balance between exploration and exploitation is crucial for solving ATD. 
We first describe how we approach the exploration using the observations collected thus far. To achieve this, we adopt a maximum-entropy strategy, selecting the measurement $q^{\text{exp}}_t$ at the $t$-th query step as 
\begin{equation}\label{eq:ent}
\small{q^{\text{exp}}_t = \arg \max_{q_t} 
\left[ 
H(\hat{x}_t | Q_t, y_{t-1})\right] = \arg \max_{q_t} -\mathbb{E}_{\hat{x}_t}[\log p(\hat{x}_t | Q_t, y_{t-1})]}
\end{equation}
Here, $H$ represents entropy, $\hat{x}_t$ denotes samples from the approximate posterior $q_{\phi_{t-1}}(x \mid y_{t-1})$. Note that we utilize the final updated prior $\phi_{t-1} = (\theta, \zeta_{t-1})$ to sample from the posterior. $Q_t$ represents the set of locations queried up to time $t$, and $ y_{t-1}$ represents the set of observations up to time $t-1$. As defined in Equation~\ref{eq:ent}, we select the query location that corresponds to the maximum entropy. To compute $\log p(\hat{x}_t | Q_t, y_{t-1})$, we begin by drawing P samples from the posterior $q_{\phi_{t-1}}(x \mid y_{t-1})$, denoting the $i$-th sample as $\hat{x}_t^i$. We then approximate $\log p(\hat{x}_t | Q_t, y_{t-1})$ using a mixture of Gaussians:
$p(\hat{x}_t | Q_t, y_{t-1}) = \sum_{i=0}^{P} \alpha_i \mathcal{N}(\hat{x}^i_t, \sigma^2_xI ) $. In the following Theorem, we derive a result that allows us to compute $q_t^{\text{exp}}$ utilizing the expression of $p(\hat{x}_t | Q_t, y_{t-1})$. 
\begin{tcolorbox}[colback=blue!5!white, colframe=blue!40!black, boxrule=0.2pt, arc=1mm]
\begin{theorem}\label{eq:argmax_error}
Assuming \( k \) represents the set of possible measurement locations at step $t$, and that all samples from the posterior $q_{\phi_{t-1}}(x \mid y_{t-1})$ have equal weights (\( \alpha_i = \alpha_j, \, \forall i, j \)), then, 
\[
\small{q^{\text{exp}}_t = \arg\max_{q_t} \left[ \sum_{i=0}^{P} \log \sum_{j=0}^{P} \exp \left( \frac{ \sum_{q_t \in k}^{ }([\hat{x}^{(i)}_t]_{q_t} - [\hat{x}^{(j)}_t]_{q_t})^2 }{2\sigma_x^2} \right) \right]},
\]
where $[.]_{q_t}$ selects element of $[.]$ indexed by $q_t$. Detailed proof is in the Appendix.
\end{theorem}
\end{tcolorbox}
The implication is that the optimal next observation location ($q^{\text{exp}}_t$) lies in the region of the search space where the predicted semantics ($\hat{x}^{(i)}_t$) show the highest disagreement among the \emph{posterior samples} ($i \in 0, \ldots, P$). This disagreement is quantified by a metric, denoted as $\mathrm{expl}^{\mathrm{score}}_{\phi}(q_t)$, which enables us to rank candidate observation locations and thereby guide exploration more effectively. In scenarios where the observation space is composed of pixels, each $\hat{x}^{(i)}_t$ represents a full image conditioned on the observed pixels $y_{t-1}$.
\vspace{-6pt}
\begin{equation}\label{eq:exp-ecore}
\small{\mathrm{expl}^{\mathrm{score}}_{\phi}(q_t) = \left[ \sum_{i=0}^{P} \sum_{j=0}^{P} \frac{([\hat{x}^{(i)}_t]_{q_t} - [\hat{x}^{(j)}_t]_{q_t})^2}{2\sigma_x^2} \right]}
\end{equation}
We now describe how the collected observations are used to guide effective exploitation. Specifically, we: $(i)$ employ a \emph{reward model} ($r_{\eta}$), parameterized by $\eta$ and incrementally trained on supervised data gathered from observations, to predict whether an observed location corresponds to the target of interest; and $(ii)$ estimate the \emph{expected log-likelihood score} at a location $q_t$, denoted as $\mathrm{likeli}^{\mathrm{score}}_{\phi}(q_t)$, defined as $\mathbb{E}_{\hat{x}_t}[ \log p(\hat{x}_t \mid Q_t, y_{t-1})]_{q_t}$. This score is key to prioritizing locations during exploitation. The following proposition provides a closed-form expression for computing $\mathrm{likeli}^{\mathrm{score}}_{\phi}(q_t)$.
\vspace{-2pt}
\begin{tcolorbox}[colback=blue!5!white, colframe=blue!40!black, boxrule=0.2pt, arc=1mm]
\begin{proposition}\label{eq:like-ecore}
The expected log-likelihood score at $q_t$ can be expressed as follows:
\[
\small{\mathrm{likeli}^{\mathrm{score}}_{\phi}(q_t) =  \sum_{i=0}^{P} \sum_{j=0}^{P} \exp \left\{ - \frac{([\hat{x}^{(i)}_t]_{q_t} - [\hat{x}^{(j)}_t]_{q_t})^2}{2\sigma_x^2} \right\}}
\]
\end{proposition}
\end{tcolorbox}
The derivation of Proposition~\ref{eq:like-ecore} is provided in Appendix. The term \( \mathrm{likeli}^{\mathrm{score}}_{\phi}(q_t) \) plays a central role in computing the exploitation score at measurement location $q_t$. Specifically, the exploitation score, \( \mathrm{exploit}^{\mathrm{score}}_{(\phi, \eta)}(q_t) \), is defined as the \emph{reward-weighted expected log-likelihood}, as shown below:
\vspace{-7pt}
\begin{equation}\label{eq:exploit-score}
\small{\mathrm{exploit}^{\mathrm{score}}_{(\phi,\eta)}(q_t) = \underbrace{\mathrm{likeli}^{\mathrm{score}}_{\phi}(q_t)}_\textit{Expected log-likelihood} \times \underbrace{\sum^{P}_{i=0} r_{\eta}([\hat{x}^{(i)}_t]_{q_t})}_\textit{reward}}
\end{equation}
As per Eqn.~\ref{eq:exploit-score}, a measurement location $q_t$ is favored for exploitation if it meets two criteria: (1) the predicted content at $q_t$ is highly likely to correspond to the target of interest, as indicated by the reward model ($r_{\eta}$); and (2) the estimated content at $q_t$, denoted as $[\hat{x}^{(i)}_t]_{q_t}$, exhibit strong agreement across the \emph{posterior samples}, suggesting high predictability about the content at that location.
Note that the reward model parameters ($\eta$) are randomly initialized and progressively updated using \emph{binary cross-entropy loss} as new supervised data becomes available after each observation.
While the reward model is initially unrefined—resulting in less reliable exploitation scores—this is not a limitation, as our sampling strategy is designed to emphasize exploration during the early stages of the discovery process. We now bring together the core components to formulate the proposed sampling strategy for efficient active target discovery. At each time step $t$, potential measurement locations ($q_t$) are ranked by jointly considering both exploration and exploitation scores, as below:
\begin{equation}\label{eq:final_score}
    \mathrm{Score}_{(\phi,\eta)}(q_t) = \alpha(\mathcal{B}) \cdot \mathrm{expl}^{\mathrm{score}}_{\phi}(q_t) + (1 - \alpha(\mathcal{B})) \cdot \mathrm{exploit}^{\mathrm{score}}_{(\phi,\eta)}(q_t)
\end{equation}
Here, $\alpha(\mathcal{B})$ is a budget-dependent function that enables a dynamic and adaptive balance between exploration and exploitation. After computing the combined score for each candidate location $q_t$ (as defined in Eqn.~\ref{eq:final_score}), the location with the highest score is selected for sampling. The choice of $\alpha(\mathcal{B})$ is problem-specific and can be tailored accordingly. A straightforward and effective option is to define it as a linear function of the remaining budget, such as $\alpha(\mathcal{B}) = \frac{\mathcal{B} - t}{\mathcal{B} + t}$. We empirically show that this simple formulation performs well across diverse application domains. The complete training and sampling algorithm is detailed in the Appendix, with a visual overview provided in the Appendix. Finally, we demonstrate that the improvement of the prior $\phi$ in turn leads to increasingly accurate estimates of the score defined in Equation~\ref{eq:final_score} as formally stated in Theorem~\ref{th:4}.
\vspace{-2pt}
\begin{tcolorbox}[colback=blue!5!white, colframe=blue!40!black, boxrule=0.2pt, arc=1mm]
\begin{theorem}\label{th:4}
Let $\phi$ and $\phi_{\text{new}}$ denote the prior parameters before and after updating the h-model using the objective in Equation~(7). Then the following relation holds, 
\[
 ||\mathrm{Score}^{*}_{(\phi^*,\eta^*)}(q_t) -\mathrm{Score}_{(\phi,\eta)}(q_t) ||  \geq ||\mathrm{Score}^{*}_{(\phi^*,\eta^*)}(q_t) -\mathrm{Score}_{(\phi_{\text{new}},\tilde{\eta})}(q_t) ||
\]
Here, $\mathrm{Score}^{*}_{(\phi^*,\eta^*)}(q_t)$ represents the true score estimate computed using the optimal prior $\phi^*$ and optimal reward model $\eta^*$, and $\tilde{\eta}$ denotes the updated reward model parameter.
\end{theorem}
\end{tcolorbox}
We present the proof in the Appendix. When tackling consecutive tasks from the same domain, we update the long-term memory using posterior samples gathered at the end of each discovery process, promoting faster and more efficient adaptation to subsequent tasks. We present a pictorial illustration of our proposed EM-PTDM approach in Figure~\ref{fig:PTDM_fm1}.
\vspace{-3pt}
\vspace{-7pt}
Next, we present our detailed experimental results to validate the efficacy of our proposed method.
\vspace{-8pt}







%% file: exp.tex
\section{Empirical Analysis}\label{sec:experiments}
\vspace{-6pt}
\textbf{Evaluation metrics.}
Since ATD seeks to maximize the identification of measurement locations containing the target of interest, we assess performance using the \emph{success rate (SR) of selecting measurement locations that belong to the target} during exploration in partially observable environments. Therefore, SR is defined as: $\mathrm{SR} = \frac{1}{L}\sum_{i=1}^{L}\frac{1}{\min{\{\mathcal{B}, U_i}\}}\sum_{t=1}^{\mathcal{B}} y_{i}^{(q_t)}; \text{where}, \> L = \text{number of tasks.}$
Here, $U_i$ denotes the maximum number of measurement locations containing the target in the $i$-th search task. We evaluate EM-PTDM and the baselines across different measurement budgets $\mathcal{B} \in \{ 150, 200, 250, 300, 350\}$ for various target categories and application domains. 
\vspace{-4pt}
\\ \\
\noindent
\textbf{Baselines.} 
We compare our proposed EM-PTDM policy to the following baselines:
\begin{itemize}[noitemsep,topsep=0pt,leftmargin=*]
    \item \emph{Random Search} (RS), in which unexplored measurement locations are selected uniformly at random.
    \item DiffATD~\citep{sarkar2025onlinefeedbackefficientactive} leverages a pre-trained diffusion model to guide sampling based on partially observed data, efficiently discovering target regions within a fixed budget. 
    \item \emph{Greedy-Adaptive} (GA) selects $q_t$ with the highest $\mathrm{exploit}_{(\phi, \eta)}^{\mathrm{score}}(q_t)$ among unexplored locations and updates reward model $r_\eta$ via binary cross-entropy after each observation.
\end{itemize}
\noindent
\vspace{-19pt}
\paragraph{Active Discovery of Unknown Species from Known Species Distribution}
We begin our evaluation of EM-PTDM in a setting where the permanent memory is trained to approximate the distribution of species from iNaturalist~\cite{inaturalist}. The distribution is formed by dividing a large geospatial region into equal-sized grids and counting target species occurrences on each grid, as detailed in Appendix. The goal is to actively discover Coccinella Septempunctata (CS) under a strict sampling budget using a prior trained on the distribution of Gladicosa and Gonioctena (GG). The results presented in Table~\ref{tab:spe}, 
suggest that EM-PTDM significantly outperforms the baseline in terms of SR, highlighting the capability of the proposed framework in ATD under an uninformative prior. We further visualize (in~\ref{fig:vis4_spe}) how EM-PTDM, starting from an uninformative prior, rapidly approximates the true distribution with only a few observations, demonstrating its swift adaptability and efficiency in data-scarce scenarios.
\vspace{-4pt}
\begin{center}
\begin{minipage}{0.56\textwidth}
  \centering
  \includegraphics[height=2.6cm, width=\linewidth]{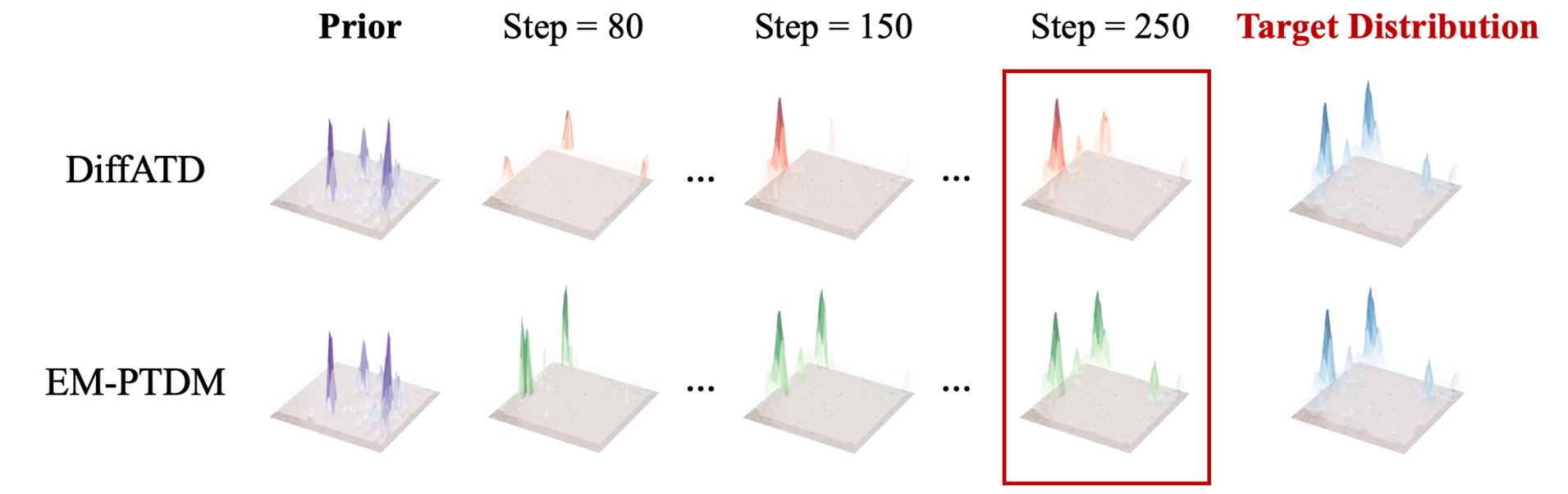}
  \captionof{figure}{\small{Active Discovery of CS species.}}
  \label{fig:vis4_spe}
\end{minipage}
\hfill
\begin{minipage}{0.43\textwidth}
  \centering
  \scriptsize
  \captionof{table}{\emph{SR} Comparison with Baselines.}
  \vspace{-2pt}
  \begin{tabular}{p{1.19cm}p{1.00cm}p{1.00cm}p{1.00cm}}
    \toprule
    \multicolumn{4}{c}{Active Discovery of CS Species with Species GG as Prior.} \\
    \midrule
    Method & $\mathcal{B}=150$ & $\mathcal{B}=200$ & $\mathcal{B}=250$ \\
    \midrule
    RS & 0.1624 & 0.2327 & 0.2775  \\
    DiffATD & 0.3420 & 0.4365 & 0.4808 \\
    GA & 0.4061 & 0.5067 & 0.5567  \\
    \hline 
    \textbf{\emph{EM-PTDM}} & \textbf{0.4983} & \textbf{0.6495} & \textbf{0.6989}  \\ 
    \bottomrule
  \end{tabular}
  \label{tab:spe}
\end{minipage}
\end{center}
\vspace{-5pt}
\paragraph{Active Discovery of Unknown Overhead Objects from Ground-view Prior}
We further assess EM-PTDM in a challenging cross-view setting, where the task is to actively discover novel objects from DOTA overhead imagery~\cite{xia2018dota}, leveraging an uninformative prior learned solely from ground-level images of semantically disjoint classes from Imagenet~\cite{deng2009imagenet}. The results are presented in Table~\ref{tab:tab4}. We observe that EM-PTDM consistently outperforms the baseline, with the performance gap widening as the search budget increases, highlighting the strength of its cumulatively refined prior in driving efficient exploration and informed target discovery. Additionally, we compare the exploration dynamics of EM-PTDM and DiffATD at two stages of the active discovery process. As shown in Figure~\ref{fig:vis4_dota}, while both models initially struggle due to uninformative priors, \textbf{EM-PTDM rapidly adapts through Doob’s h-transform, enabling more efficient target discovery as observations accumulate.}
\vspace{-5pt}
\begin{center}
\begin{minipage}{0.51\textwidth}
  \centering
  \includegraphics[height=2.5cm,width=\linewidth]{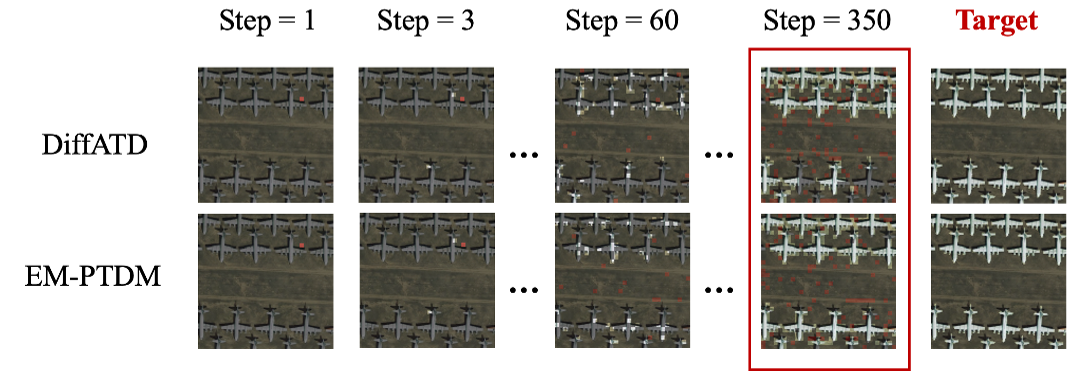}
  \captionof{figure}{\small{Active Discovery of Overhead objects.}}
  \label{fig:vis4_dota}
\end{minipage}
\hfill
\begin{minipage}{0.48\textwidth}
  \centering
  \scriptsize
  \captionof{table}{\emph{SR} Comparison with Baselines.}
  \vspace{-2pt}
  \begin{tabular}{p{1.19cm}p{1.00cm}p{1.00cm}p{1.00cm}}
    \toprule
    \multicolumn{4}{c}{Active Discovery of Overhead Objects with ImageNet as Prior.} \\
    \midrule
    Method & $\mathcal{B}=250$ & $\mathcal{B}=300$ & $\mathcal{B}=350$ \\
    \midrule
    RS & 0.2325 & 0.2852 & 0.3207  \\
    DiffATD & 0.5143 & 0.6391 & 0.7348 \\
    GA & 0.4784 & 0.5659 & 0.6562   \\
    \hline 
    \textbf{\emph{EM-PTDM}} & \textbf{0.5620} & \textbf{0.7013} & \textbf{0.8256}  \\ 
    \bottomrule
  \end{tabular}
  \label{tab:tab4}
\end{minipage}
\end{center}

\vspace{-12pt}
\paragraph{Enhancing Permanent Memory with Domain Cues for Improved In-Domain Target Discovery}
In real-world settings, it's common to encounter scenarios where ATD tasks must be addressed consecutively within similar domains, requiring models that can retain and transfer knowledge effectively. To support efficient adaptation in sequential ATD tasks within the same domain, we update the permanent memory—specifically, the pretrained diffusion model ($s^{\theta^*}$)—using posterior samples generated at the final step of the previous task, conditioned on all collected observations. 
\begin{wraptable}{r}{0.41\textwidth}
  \vspace{-10pt} 
  \scriptsize
  \caption{Effect of Permanent Memory}
  \label{tab:pm}
  \begin{tabular}{p{1.15cm}p{0.99cm}p{0.99cm}p{0.99cm}}
    \toprule
    \multicolumn{4}{c}{ATD of Overhead Objects with ImageNet as Prior.} \\
    \midrule
    $s^{\theta^*}$ update & $\mathcal{B}=250$ & $\mathcal{B}=300$ & $\mathcal{B}=350$ \\
    \midrule
    No & 0.5620 & 0.7013 & 0.8256 \\ 
    \textbf{Yes} & \textbf{0.5859} & \textbf{0.7194} &\textbf{0.8461} \\
    \bottomrule
  \end{tabular}
  \vspace{-10pt} 
\end{wraptable}
This update allows the permanent memory to gain domain-specific structural knowledge, thereby reducing the divergence between prior and posterior distributions in subsequent tasks. As a result, the learning task of doob's $h$-model becomes easier as the correction needed for rapid adaptation diminishes, enabling the model to explore and discover new targets more effectively with fewer observations. We assess the impact of updating permanent memory after each task by comparing it against a static-memory baseline. 
For this comparison, we consider the task of ATD of overhead objects with ImageNet as the prior. As reported in~\ref{tab:pm}, the results reveal a marked improvement in the SR when updates are applied, underscoring the value of \textbf{accumulating domain-specific knowledge in permanent memory to accelerate adaptation and enhance target discovery in related tasks.}

\begin{figure}[!h]
\includegraphics[width=\linewidth, trim=0pt 0pt 0 90pt, clip]{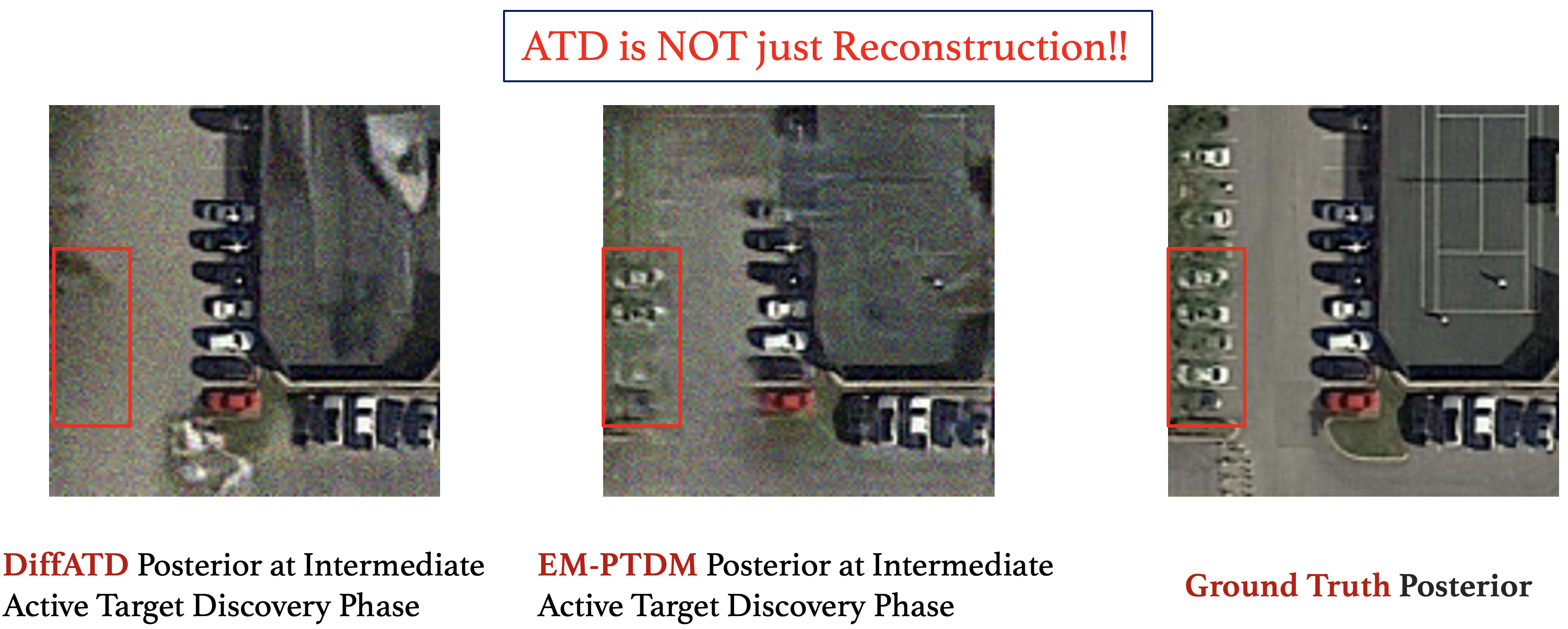}
\caption{ ATD is about Discovering Targets, NOT just Reconstructing the Search Space.}
\label{fig: atd_vs_recon8}
\end{figure}

\vspace{-10pt} 
\paragraph{The Role of Transient Memory in Rapid Adaptation to Novel ATD Tasks}
To assess the impact of transient memory, we compare posterior estimates generated with and without the proposed transient memory mechanism, starting from an uninformative prior. As illustrated in Fig.~\ref{fig:tm}, incorporating transient memory via the h-model enables rapid adaptation from sparse early observations, resulting in posterior estimates that closely approximate the ground truth and facilitate more efficient target discovery. In contrast, the baseline, which relies solely on permanent memory, yields an incorrect posterior, resulting in limited exploration capability in early stages, underscoring the critical role of the $h$-model in accelerating effective exploration. We quantitatively assess the role of the $h$-model by measuring $L2$ distance-based semantic dissimilarity between the posterior and ground-truth targets. Results (~\ref{fig:vis4_tm}) show that \textbf{incorporating the $h$-model significantly accelerates convergence to the true posterior}, as dissimilarity drops much faster compared to using the permanent memory alone, demonstrating the $h$-model’s effectiveness in rapidly correcting the prior with minimal observations.
\vspace{-5pt}
\begin{center}
\begin{minipage}{0.41\textwidth}
  \centering
  \includegraphics[height=2.4cm, width=\linewidth]{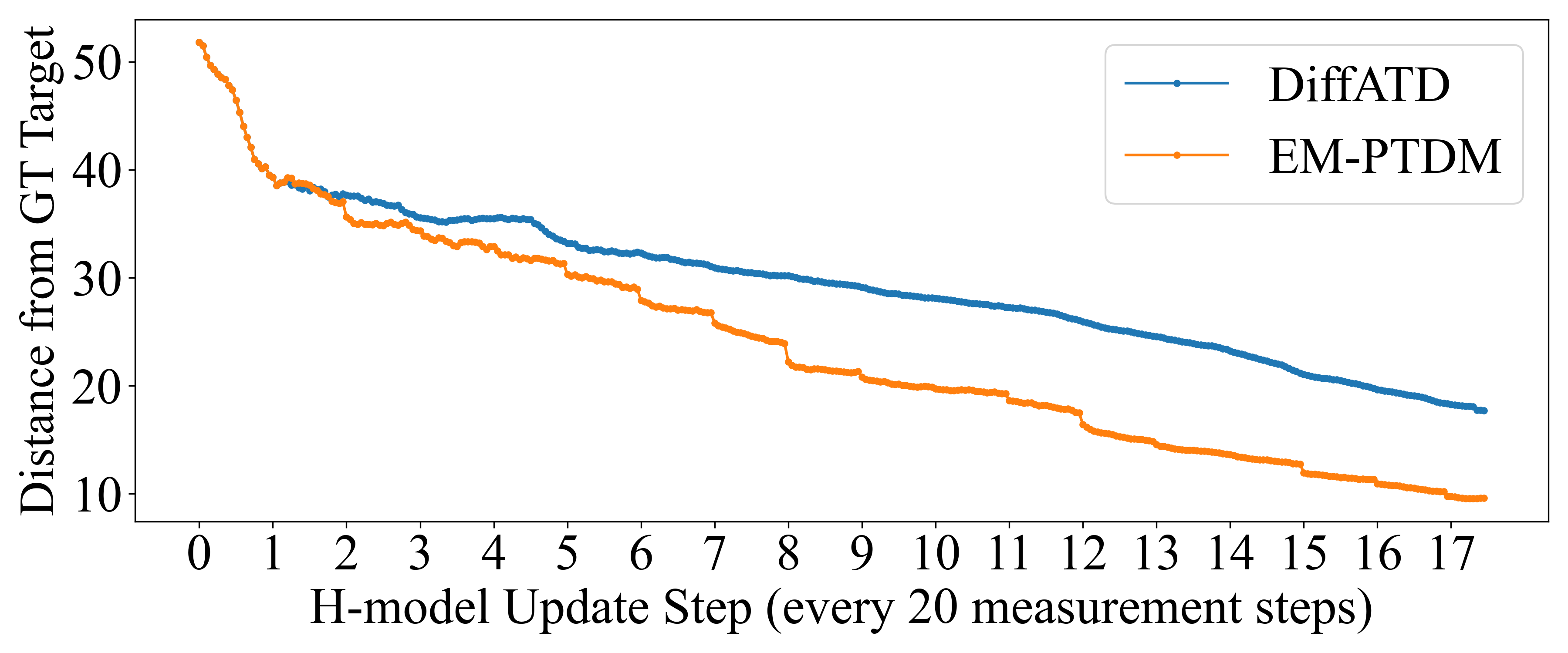}
  \captionof{figure}{\small{Effectiveness of Transient Memory.}}
  \label{fig:vis4_tm}
\end{minipage}
\hfill
\begin{minipage}{0.57\textwidth}
  \centering
  \captionof{figure}{\small{Posterior Dynamics with and without $h$-model.}}
  \vspace{-4pt}
  \includegraphics[width=\linewidth]{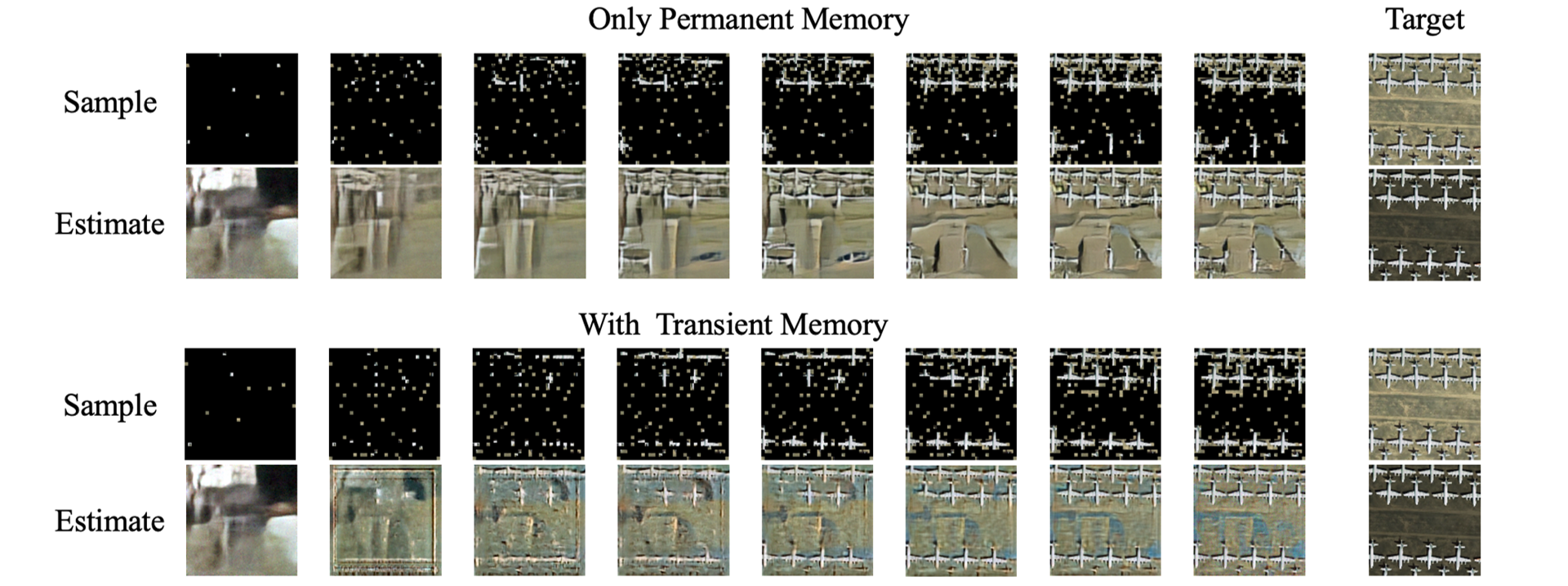}
  \label{fig:tm}
\end{minipage}
\end{center}
\vspace{-18pt}
\paragraph{An Important Observation: ATD is not just About Reconstruction}
\vspace{-5pt}
One might ask: if the primary goal is for the posterior samples to accurately reconstruct the search space, isn’t that sufficient for efficient target discovery? Interestingly, in our setting, a precise reconstruction of the entire search space is not strictly necessary, as long as the model effectively identifies and reconstructs the target regions, efficient discovery can still be achieved. To validate this hypothesis, we conduct an experiment and visualize the posterior at intermediate stages of active target discovery. We compare posterior samples from EM-PTDM and DiffATD using a representative task where EM-PTDM significantly outperforms DiffATD, allowing us to understand how the posterior contributes to improved target discovery.
We present the visualization in the Figure~\ref{fig: atd_vs_recon8}. Our observations reveal that, \textbf{while EM-PTDM’s posterior samples exhibit lower overall reconstruction quality compared to those from DiffATD, they more effectively focus on target-rich regions} (For example, see the highlighted Red box in the Figure~\ref{fig: atd_vs_recon8}), leading to significantly improved target discovery performance. This highlights a key insight: successful target discovery relies more on accurately modeling the regions of interest than on reconstructing the entire search space.
\vspace{-12pt}
\vspace{-23pt}

%% file: abl.tex

%% file: rel_work.tex
\section{Related Work}
\vspace{-8pt}
\noindent
Several RL-based methods~\citep{uzkent2020learning,sarkar2024visual,sarkar2023partially,nguyen2024amortized} have been developed for ATD; however, they typically assume full observability of the search space and require access to large-scale, pre-labeled datasets to learn efficient exploration strategies. Training-free approaches inspired by Bayesian decision theory~\citep{garnett2012bayesian,jiang2017efficient,jiang2019cost} offer an appealing alternative, yet their dependence on complete observability limits their effectiveness in partially observable settings. More recently, a new class of methods~\citep{rangrej2022consistency,pirinen2022aerial,sarkar2024gomaa} has emerged, explicitly addressing active discovery under partial observability. Nevertheless, these approaches still face a major challenge—their heavy reliance on extensive annotated datasets to reach optimal performance. DiffATD~\citep{sarkar2025onlinefeedbackefficientactive} advances active target discovery by enabling efficient exploration in partially observable environments without requiring task-specific labeled data. However, it still depends on domain-specific prior samples to learn an effective prior model, limiting its applicability in domains where such data is unavailable. To address this critical gap, we propose EM-PTDM, a novel framework that enables active target discovery without relying on any prior domain samples.
\vspace{-10pt}

%% file: conclusion.tex
\section{Conclusion}
\vspace{-7pt}
We propose EM-PTDM, a principled and generalizable framework for ATD in partially observable environments, remarkably operating without any task-specific prior data, significantly broadening its applicability in diverse domains. EM-PTDM bridges solid theoretical foundations with compelling empirical performance across diverse domains, ranging from rare species discovery to remote sensing.

%% file: appendix.tex
\newpage

\newcommand{\ldotsfill}{\leavevmode\leaders\hbox to .5em{\hss.\hss}\hfill\kern0pt}

\section*{Appendix: Active Target Discovery under Uninformative Prior:
The Power of Permanent and Transient Memory}
\vspace{3pt}

\section*{Overview of the Contents}

\noindent
\begin{tabularx}{\textwidth}{@{}Xr@{}}
\textbf{A:} \textbf{Proofs of Theoretical Results}  \ldotsfill & 21 \\ 
\quad A.1 Proof of Proposition 1 \ldotsfill & 21 \\ [0.5em]
\quad A.2 Proof of Lemma 1 \ldotsfill & 21 \\ [0.5em]
\quad A.3 Proof of Theorem 1 \ldotsfill & 22 \\ [0.5em]
\quad A.4 Proof of Theorem 2 \ldotsfill & 22 \\  [0.5em]
\quad A.5 Proof of Theorem 3 \ldotsfill & 23 \\ [0.5em]
\quad A.6 Proof of Theorem 4 \ldotsfill & 24-25 \\ [0.5em]
\quad A.7 Proof of Proposition 2 \ldotsfill & 25-26 \\  [0.5em]

\textbf{B: Empirical Analysis of $h$ model} \ldotsfill & 26 \\ 
\quad B.1 Doob's $h$-transform as the Correction Factor \ldotsfill & 26 \\ [0.5em]
\quad B.2 Details of $h$-model Update Scheduler \ldotsfill & 26-27 \\ [0.5em]
\quad B.3 Effect of $h$-model Update Scheduler \ldotsfill & 27 \\  [0.5em]

\textbf{C: Exploratory Nature of EM-PTDM} \ldotsfill & 27 \\ [0.5em]

\textbf{D: Training and Inference Pseudocode} \ldotsfill & 27-28 \\ [0.5em]

\textbf{E: Efficacy of $h$-model In Estimating The Search Space With Only Few Observations} \ldotsfill & 29 \\ [0.5em]

\textbf{F: Analyzing the Role of Permanent and Transient Memory for Enhancing In-Domain Target Discovery} \ldotsfill & 29 \\ [0.5em]

\textbf{G: Effect of $\kappa{(\beta)}$} \ldotsfill & 29-30 \\ [0.5em]

\textbf{H: EM-PTDM’s Capability of Discovering Isolated Targets within Observation Budget} \ldotsfill & 30-31 \\ [0.5em]

\textbf{I: Species Distribution Modelling as Active Target Discovery Problem} \ldotsfill & 31 \\ [0.5em]

\textbf{J: Additional Results on Active Discovery of Unknown Species from Known Species Distribution} \ldotsfill & 31-32 \\ [0.5em]

\textbf{K: More Visualizations on Efficiency of h-model’s Adaptability From Very Sparse Observations} \ldotsfill & 32-33 \\ [0.5em]

\textbf{L: More Visualizations of the Exploration Behavior of EM-PTDM at Different Active Target Discovery Phases} \ldotsfill & 33-34 \\ [0.5em]

\textbf{M: Active Target Discovery of Balls Using MNIST Images as the Prior} \ldotsfill & 35 \\ [0.5em]
\quad M.1 Dataset Creation Procedure \ldotsfill & 35 \\ [0.5em]
\quad M.2 SR Comparisons with Baseline Approaches \ldotsfill & 35 \\ [0.5em]
\quad M.3 Analyzing the Exploration Strategies of EM-PTDM and DiffATD Under Increasing Task Complexity \ldotsfill & 35-36 \\ [0.5em]

\textbf{N: Architecture, Training Details: $h$-model, Pretrained Diffusion Model, and the Reward Model; and Computing Resources} \ldotsfill & 36 \\ [0.5em]
\quad N.1 Details of $h$-model \ldotsfill & 36 \\ [0.5em]
\quad N.2 Details of Reward Model \ldotsfill & 36-37 \\ [0.5em]
\quad N.3 Details of Primary Memory as Pretrained Diffusion Model \ldotsfill & 36-37 \\ [0.5em]

\textbf{O:  Statistical Significance Results of EM-PTDM} \ldotsfill & 37-38 \\ [0.5em]

\textbf{P: Impact of Weak Permanent Memory on Active Target Discovery
Performance} \ldotsfill & 38 \\  [0.5em]

\textbf{Q: More Details on Computational Cost across Search Space} \ldotsfill & 38 \\ [0.5em]

\textbf{R: Code Link} \ldotsfill & 39 \\ [0.5em]

\end{tabularx}

\newpage

\section{Proofs of Theoretical Results}

\subsection{Proof of Proposition 1}\label{app:1}
\begin{proof}
\begin{align}
\theta^* &= \arg\max_{\theta} \, \mathbb{E}_{p(y)} \left[ \log q_{\theta}(y) \right] \label{eq:em-start} \\
         &= \arg\min_{\theta} \, \mathrm{KL}\left( p(y) \,\|\, q_{\theta}(y) \right) \label{eq:kl-em}
\end{align}

The core principle of the EM algorithm is that, for any two parameter sets $\theta_a$ and $\theta_b$, the following identity holds:

\begin{align}
\log \frac{q_{\theta_a}(y)}{q_{\theta_b}(y)} &= \log \frac{q_{\theta_a}(x, y)}{q_{\theta_b}(x, y)} \cdot \frac{q_{\theta_b}(x \mid y)}{q_{\theta_a}(x \mid y)} \\
&= \mathbb{E}_{q_{\theta_b}(x \mid y)} \left[ \log \frac{q_{\theta_a}(x, y)}{q_{\theta_b}(x, y)} \right] + \mathrm{KL} \left( q_{\theta_b}(x \mid y) \,\|\, q_{\theta_a}(x \mid y) \right) \\
&\geq \mathbb{E}_{q_{\theta_b}(x \mid y)} \left[ \log q_{\theta_a}(x, y) - \log q_{\theta_b}(x, y) \right] \label{eq:em-lower-bound}
\end{align}

This inequality remains valid when taking the expectation over $p(y)$. Consequently, starting from an initial parameter setting $\theta_0$, the EM update rule can be expressed as:

\begin{align}
\theta_{k+1} &= \arg\max_{\theta} \, \mathbb{E}_{p(y)} \mathbb{E}_{q_{\theta_k}(x \mid y)} \left[ \log q_{\theta}(x, y) - \log q_{\theta_k}(x, y) \right] \\
             &= \arg\max_{\theta} \, \mathbb{E}_{p(y)} \mathbb{E}_{q_{\theta_k}(x \mid y)} \left[ \log q_{\theta}(x, y) \right] \label{eq:em-update}
\end{align}

In the empirical Bayes setting, the forward model $p(y \mid x)$ is known and only the parameters of the prior $q_{\theta}$(x) should be optimized. In this case, Eq.~\ref{eq:em-update} becomes:

\begin{align}
\theta_{k+1} &= \arg\max_{\theta} \, \mathbb{E}_{p(y)} \mathbb{E}_{q_{\theta_k}(x \mid y)} \left[ \log q_{\theta}(x) + \log p(y \mid x) \right] \label{eq:em-general} 
\end{align}

\begin{equation}
\theta_{k+1} = \arg\max_{\theta} \, \mathbb{E}_{p(y)} \mathbb{E}_{q_{\theta_k}(x \mid y)} \left[ \log q_{\theta}(x) \right]
\end{equation}

\end{proof}

\subsection{Proof of Lemma 1}\label{app:2}

\begin{proof}
We can express the conditional score as follows:
\begin{align}
\nabla_{x_t} \ln p_t(x \mid y) &= \int \nabla_{x_t} \ln \overrightarrow{p}_{t|0}(x_t \mid x_0) \, \overleftarrow{p}_{0|t}(x_0 \mid x_t, y) \, dx_0 
\end{align}
The minimizer is obtained by exploiting the property that the conditional expectation yields the optimal solution under the mean squared error criterion:

\begin{align}
\mathbb{E} \left[ \nabla_{H_t} \ln p_{t|0}(H_t \mid X_0) - \nabla_{H_t} \ln p_t(H_t) \, \big| \, Y = y, H_t = x \right] 
\end{align}

\begin{align}
h_t^*(x, y) 
&= \left( \int \left[ \nabla_x \ln p_{t|0}(x \mid x_0) - \nabla_x \ln p_t(x) \right] p_{0|t}(x_0 \mid H_t = x, Y = y) \, dx_0 \right) \\
&= \int \nabla_x \ln p_{t|0}(x \mid x_0) \, p_{0|t}(x_0 \mid H_t = x, Y = y) \, dx_0 - \nabla_x \ln p_t(x)
\end{align}

\begin{align}
&= \nabla_{x} \ln p_t(x \mid y) - \nabla_{x} \ln p_t(x) \\
&= \nabla_{x} \ln p_t(y \mid x)
\end{align}
    
\end{proof}

\subsection{Proof of Theorem 1}\label{app:3}
\begin{proof}

\begin{align}
\arg \max_{\phi}  \mathbb{E}_{p(y)}[\log q_{\phi}(y)]
\end{align}

Utilizing the result of Theorem 1, we can express the above expression as follows:
\begin{align}
\phi^{\text{new}} = \arg\max_{\phi} \, \mathbb{E}_{p(y)} \, \mathbb{E}_{\text{{$q_{\phi}(x|y)$}}} \left[ \log q_{\phi}(x) \right]
\end{align}
Maximizing the above objective involves computing $\nabla_x \log q_{\phi}(x)$. Following the approach in denoising score matching, the above optimization can be equivalently reformulated as:

\begin{align}\label{eq:dsm}
\zeta^{\text{new}} =\min_{\zeta} \, \mathbb{E}_{(X_0, Y),\, \varepsilon,\, t} \left\| \left( h^{\zeta}_t(x_t, Y) + s^{\theta^*}_t(x_t) \right) - \varepsilon \right\|^2
\end{align}
\noindent
Let $x_t = \sqrt{\bar{\alpha}_t} \, X_0 + \sqrt{1 - \bar{\alpha}_t} \, \varepsilon$, where $X_0 \sim q_{\phi}(x_0 \mid y), Y \sim y$ and $\varepsilon \sim \mathcal{N}(0, \mathbf{I})$. Also, $\phi^{\text{new}} = (\theta^*, \zeta^{\text{new}})$.

Thus, optimizing the objective in Equation~\ref{eq:dsm}, is equivalent to one step of EM. Hence, guarantees the improvement of expected log-evidence. As a result, we can write:
\[
   \mathbb{E}_{p(y)}[\log q_{\phi_{\text{new}}}(y)] \geq \mathbb{E}_{p(y)}[\log q_{\phi}(y)].
\]

\end{proof}

\subsection{Proof of Theorem 2}\label{app:4}
\begin{proof}
\begin{align}
\mathbb{E}_{p(y_{t+k})}[\log q_{\phi_{t+k}}(y)]
\end{align}

Assuming the observations collected during the active discovery process are independent, we can decompose the above expression into two parts as follows:

\begin{align}
&= \underbrace{\mathbb{E}_{p(y_{t})}[\log q_{\phi_{t+k}}(y)]}_{\textit{part1}} \times \underbrace{\mathbb{E}_{p(y_{t+k: (t+1)})}[\log q_{\phi_{t+k}}(y)]}_{\textit{part2}}
\end{align}

The set of observations gathered from time step $t+1$ to $t+k$ is denoted as $y_{t+k: (t+1)}$.

For \textit{Part1}, we can write, 
\begin{align}\label{eq:33}
\mathbb{E}_{p(y_{t})}[\log q_{\phi_{t+k}}(y)] \geq \mathbb{E}_{p(y_{t})}[\log q_{\phi_{t}}(y)] \>\>\>\>\text{(Following Theorem 2)}
\end{align}

For \textit{Part2}, we can write, 

\begin{align}\label{eq:34}
\mathbb{E}_{p(y_{t+k: (t+1)})}[\log q_{\phi_{t+k}}(y)] \geq \mathbb{E}_{p(y_{t+k: (t+1)})}[\log q_{\phi_{t}}(y)] \>\>\>\> \text{(By Definition)}
\end{align}

The above relation holds because, unlike $\phi_k$, the model $\phi_{t+k}$ is trained on posterior samples explicitly conditioned on the observations $y_{t+k: (t+1)}$. As a result, when $\phi_{t+k}$ is optimally trained, the left-hand side of Equation~\ref{eq:34} reaches its optimal value. In contrast, since $\phi_k$ is never exposed to $y_{t+k: (t+1)}$ during training, it cannot attain the optimal value of the left-hand side in Equation~\ref{eq:34}.

Finally, combining the results of Equation~\ref{eq:33} and~\ref{eq:34}, we can write:
\[
\mathbb{E}_{p(y_{t+k})}[\log q_{\phi_{t+k}}(y)] \geq \mathbb{E}_{p(y_{t+k})}[\log q_{\phi_{t}}(y)]
\]

\end{proof}

\subsection{Proof of Theorem 3}\label{app:5}
\begin{proof}
We start with the definition of $q^{\text{exp}}_t$ as follows:
\[
q^{\text{exp}}_t =\arg \max_{q_t} -\mathbb{E}_{\hat{x}_t}[\log p(\hat{x}_t | Q_t, y_{t-1})]\>\>\> \text{where}\>\>\> p(\hat{x}_t | Q_t, y_{t-1}) = \sum_{i=0}^{P} \alpha_i \mathcal{N}(\hat{x}^i_t, \sigma^2_xI )
\]
According to~\citep{hershey2007approximating}, 
\[
q^{\text{exp}}_t \propto \sum^{P}_{i=0} \alpha_i \text{log} \sum^{P}_{j=0} \alpha_j \text{exp} \left\{ {\frac{||\hat{x}^{(i)}_t - \hat{x}^{(j)}_t||^2_2}{2\sigma^2_x}} \right\}
\]
We can write,
\[
q^{\text{exp}}_t = \arg\max_{q_t} \sum^{P}_{i=0} \alpha_i \text{log} \sum^{P}_{j=0} \alpha_j \text{exp} \left\{ {\frac{||\hat{x}^{(i)}_t - \hat{x}^{(j)}_t||^2_2}{2\sigma^2_x}} \right\}
\]

\[
q^{\text{exp}}_t = \arg\max_{q_t} \sum_{i,j} \log \left( \exp \left( \frac{\| \hat{x}_t^{(i)} - \hat{x}_t^{(j)} \|^2}{2\sigma_x^2} \right) \right)  (\text{By assuming}, \alpha_i = \alpha_j, \, \forall i, j )
\]

\[
q^{\text{exp}}_t = \arg\max_{q_t} \sum_{i,j} \log \left( \exp \left( \frac{\sum_{a \in Q_t} ([\hat{x}^{(i)}_t]_a - [\hat{x}^{(j)}_t]_a)^2}{2\sigma_x^2} \right) \right)
\]

We decompose it into two parts: one representing the set of potential measurement locations at the query step 
$t$, and the other corresponding to the set of locations already selected in $Q_{t-1}$. Hence, we can express it as follows:

\[
q^{\text{exp}}_t = \arg\max_{q_t} \sum_{i,j} \log \left( \exp \left( \frac{\sum_{q_t \in k} ([\hat{x}^{(i)}_t]_{q_t} - [\hat{x}^{(j)}_t]_{q_t})^2 + \sum_{r \in Q_{t-1}} ([\hat{x}^{(i)}_t]_r - [\hat{x}^{(j)}_t]_r)^2}{2\sigma_x^2} \right) \right)
\]

\[
q^{\text{exp}}_t = \arg\max_{q_t} \sum_{i,j} \log \left( \prod_{q_t \in k} \exp \left( \frac{([\hat{x}^{(i)}_t]_{q_t} - [\hat{x}^{(j)}_t]_{q_t})^2}{2\sigma_x^2} \right) \prod_{r \in Q_{t-1}} \exp \left( \frac{([\hat{x}^{(i)}_t]_r - [\hat{x}^{(j)}_t]_r)^2}{2\sigma_x^2} \right) \right)
\]

\[
q^{\text{exp}}_t \propto \arg\max_{q_t} \sum_{i,j} \left( \sum_{q_t \in k} \frac{([\hat{x}^{(i)}_t]_{q_t} - [\hat{x}^{(j)}_t]_{q_t})^2}{2\sigma_x^2} + \underbrace{\sum_{r \in Q_{t-1}} \frac{([\hat{x}^{(i)}_t]_r - [\hat{x}^{(j)}_t]_r)^2}{2\sigma_x^2}}_\text{We can ignore as it doesn't depend on the choice of measurement location at time t.} \right)
\]

\[
q^{\text{exp}}_t \propto \arg\max_{q_t} \sum_{i,j} \sum_{q_t \in k} \frac{([\hat{x}^{(i)}_t]_{q_t} - [\hat{x}^{(j)}_t]_{q_t})^2}{2\sigma_x^2}.
\]

\[
q^{\text{exp}}_t = \arg\max_{q_t} \left[ \sum_{i=0}^{P} \log \sum_{j=0}^{P} \exp \left( \frac{ \sum_{q_t \in k}^{ }([\hat{x}^{(i)}_t]_{q_t} - [\hat{x}^{(j)}_t]_{q_t})^2 }{2\sigma_x^2} \right) \right]
\]
    
\end{proof}

\subsection{Proof of Theorem 4}\label{app:7}

\begin{proof}
Using Equation 12, we can decompose the score as follows:
\begin{align}
&||\mathrm{Score}^{*}_{(\phi^*,\eta^*)}(q_t) -\mathrm{Score}_{(\phi,\eta)}(q_t) || = \alpha(\mathcal{B}) \underbrace{\big( \mathrm{expl}^{\mathrm{score}}_{\phi^*}(q_t) - \mathrm{expl}^{\mathrm{score}}_{\phi}(q_t) \big)}_{\textit{A}} \nonumber \\
&\quad + (1 - \alpha(\mathcal{B})) \underbrace{\big( \mathrm{exploit}^{\mathrm{score}}_{(\phi^*,\eta^*)}(q_t) - \mathrm{exploit}^{\mathrm{score}}_{(\phi,\eta)}(q_t) \big)}_{\textit{B}}
\end{align}
By following a similar decomposition, we can write, 
\begin{align}
&||\mathrm{Score}^{*}_{(\phi^*,\eta^*)}(q_t) -\mathrm{Score}_{(\phi_{\text{new}},\tilde{\eta})}(q_t) || = \alpha(\mathcal{B}) \underbrace{\big( \mathrm{expl}^{\mathrm{score}}_{\phi^*}(q_t) - \mathrm{expl}^{\mathrm{score}}_{\phi_{\text{new}}}(q_t) \big)}_{\textit{C}} \nonumber \\
&\quad + (1 - \alpha(\mathcal{B})) \underbrace{\big( \mathrm{exploit}^{\mathrm{score}}_{(\phi^*,\eta^*)}(q_t) - \mathrm{exploit}^{\mathrm{score}}_{(\phi_{\text{new}},\tilde{\eta)}}(q_t) \big)}_{\textit{D}}
\end{align}

We can write,
\begin{align}
&\underbrace{\big( \mathrm{expl}^{\mathrm{score}}_{\phi^*}(q_t) - \mathrm{expl}^{\mathrm{score}}_{\phi}(q_t) \big)}_{\textit{A}} \geq \underbrace{\big( \mathrm{expl}^{\mathrm{score}}_{\phi^*}(q_t) - \mathrm{expl}^{\mathrm{score}}_{\phi_{\text{new}}}(q_t) \big)}_{\textit{C}}\>\>\>  \nonumber \\
&\text{(Follows from Proposition 1 and Theorem 1)} \label{37}
\end{align}

The above relation holds as $\phi_{\text{new}}$ is obtained by applying one iteration of EM, thus guaranteeing improvement from the previous iteration prior ($\phi$). Hence, ensure a more accurate exploration score estimation compared to the prior of previous iteration. 

Next, we will compare the terms B and D and show that
\[
\underbrace{\big( \mathrm{exploit}^{\mathrm{score}}_{(\phi^*,\eta^*)}(q_t) - \mathrm{exploit}^{\mathrm{score}}_{(\phi,\eta)}(q_t) \big)}_{\textit{B}} \geq \underbrace{\big( \mathrm{exploit}^{\mathrm{score}}_{(\phi^*,\eta^*)}(q_t) - \mathrm{exploit}^{\mathrm{score}}_{(\phi_{\text{new}},\tilde{\eta)}}(q_t) \big)}_{\textit{D}}
\]
Utilizing the expression in 11, we compare the exploit scores computed via different parameterizations of the prior as follows:

\[
\mathrm{exploit}^{\mathrm{score}}_{(\phi,\eta)}(q_t) = \underbrace{\mathrm{likeli}^{\mathrm{score}}_{\phi}(q_t)}_\textit{Expected log-likelihood} \times \underbrace{\sum^{P}_{i=0} r_{\eta}([\hat{x}^{(i)}_t]_{q_t})}_\textit{reward}
\]
We can rewrite the above expression as follows:
\[
\mathrm{exploit}^{\mathrm{score}}_{(\phi,\eta)}(q_t) = \mathrm{likeli}^{\mathrm{score}}_{\phi}(q_t) + \mathbb{E}_{[\hat{x}_t^{(i)}] \sim q_{\phi(x \mid y)}} [r_{\eta}([\hat{x}_t^{(i)}]_{q_t})]
\]

Following exactly similar steps, we can also write
\[
\mathrm{exploit}^{\mathrm{score}}_{(\phi_{\text{new}},\tilde{\eta})}(q_t) = \mathrm{likeli}^{\mathrm{score}}_{\phi_{\text{new}}}(q_t) + \mathbb{E}_{[\hat{x}_t^{(i)}] \sim q_{\phi_{\text{new}}(x \mid y)}} [r_{\tilde{\eta}}([\hat{x}_t^{(i)}]_{q_t})]
\]

Following the same reasoning as in 37, we can write
\begin{align}
&\big( \mathrm{likeli}^{\mathrm{score}}_{\phi^*}(q_t) - \mathrm{likeli}^{\mathrm{score}}_{\phi}(q_t) \big) \geq \big( \mathrm{likeli}^{\mathrm{score}}_{\phi^*}(q_t) - \mathrm{likeli}^{\mathrm{score}}_{\phi_{\text{new}}}(q_t) \big)\>\>\>  \nonumber \\
&\text{(Follows from Proposition 1 and Theorem 1)} 
\end{align}

Now, note that at the beginning of the active target discovery process, $r_{\tilde{\eta}}([\hat{x}_t^{(i)}]_{q_0}) = r_{\eta}([\hat{x}_t^{(i)}]_{q_0})$.

As $\phi_{\text{new}}$ is a improved parameterization of the previous iteration prior $\phi$, thus posterior samples from $q_{\phi_{\text{new}}}(x \mid y)$ are more reliable and accurate compared to the posterior samples from $q_{\phi}(x \mid y)$ and thus reward evaluated on the posterior samples from $q_{\phi_{\text{new}}}(x \mid y)$ are more reliable. Furthermore, the reward model $r_{\tilde{\eta}}$ benefits from training on more diverse samples, as entropy computed using the updated prior $\phi_{\text{new}}$ is more accurate than that from $\phi$. This leads to improved data diversity during collection, enabling $r_{\tilde{\eta}}$ to converge faster than $r_{\eta}$. The following lemma supports this hypothesis:

\begin{lemma}[Diverse Data Improves Convergence]\label{lm:a}~\citep{hardt2016train, yin2018gradient}
Let $\theta_t$ be the parameters of a neural network trained using SGD with batch size 1 and learning rate $\eta$ on dataset $S$. Let $\mathcal{L}_S(\theta)$ be the empirical loss. Assume the loss is $L$-smooth and gradients are bounded by $G$. Let $S_1$ and $S_2$ be two datasets of size $n$, with $D(S_1) > D(S_2)$. Then, for the same number of iterations $T$, the expected generalization gap
\[ 
\mathbb{E}[\mathcal{L}(\theta_T^{(1)}) - \mathcal{L}(\theta_T^{(2)})] < 0 \>\>\> \text{where,}\>\>
D(S) = \frac{1}{n^2} \sum_{i,j} \|x_i - x_j\|^2 \quad \text{for } (x_i, y_i), (x_j, y_j) \in S
\]
where $\theta_T^{(1)}$ and $\theta_T^{(2)}$ are trained on $S_1$ and $S_2$ respectively, assuming the data distribution $\mathcal{D}$ has high support over $\mathcal{X}$.
\end{lemma}

Hence, $\theta_T^{(1)}$ is closer to optimal solution of $\mathcal{L}(\theta)$ than $\theta_T^{(2)}$ in fewer steps, assuming same training budget.

Thus, utilizing the result of the lemma~\ref{lm:a}, we can write:
\begin{align}
&\big(\mathbb{E}_{[\hat{x}_t^{(i)}] \sim q_{\phi^{*}(x \mid y)}} [r_{\eta^*}([\hat{x}_t^{(i)}]_{q_t})] -
\mathbb{E}_{[\hat{x}_t^{(i)}] \sim q_{\phi(x \mid y)}} [r_{\eta}([\hat{x}_t^{(i)}]_{q_t})]\big) \geq  \nonumber \\
&\big(\mathbb{E}_{[\hat{x}_t^{(i)}] \sim q_{\phi^{*}(x \mid y)}} [r_{\eta^*}([\hat{x}_t^{(i)}]_{q_t})] -
\mathbb{E}_{[\hat{x}_t^{(i)}] \sim q_{\phi_{\text{new}}(x \mid y)}} [r_{\tilde{\eta}}([\hat{x}_t^{(i)}]_{q_t})]\big) \label{39}
\end{align}

Now, combining the results of (38) and (39), we can write
\begin{align}
&\underbrace{\big( \mathrm{exploit}^{\mathrm{score}}_{(\phi^*,\eta^*)}(q_t) - \mathrm{exploit}^{\mathrm{score}}_{(\phi,\eta)}(q_t) \big)}_{\textit{B}} \geq \underbrace{\big( \mathrm{exploit}^{\mathrm{score}}_{(\phi^*,\eta^*)}(q_t) - \mathrm{exploit}^{\mathrm{score}}_{(\phi_{\text{new}},\tilde{\eta)}}(q_t) \big)}_{\textit{D}} \label{40}
\end{align}

Finally, leveraging the results of (37) and (40), and utilizing the definition of (35), we can write

\[
 ||\mathrm{Score}^{*}_{(\phi^*,\eta^*)}(q_t) -\mathrm{Score}_{(\phi,\eta)}(q_t) ||  \geq ||\mathrm{Score}^{*}_{(\phi^*,\eta^*)}(q_t) -\mathrm{Score}_{(\phi_{\text{new}},\tilde{\eta})}(q_t) ||
\]

\end{proof}

\subsection{Proof of Proposition 2}\label{app:6}

\begin{proof}
    We start with the definition of entropy $H$:
\[
\mathbb{E}_{\hat{x}_t}[\log p(\hat{x}_t | Q_t, y_{t-1})]=-H(\hat{x}_t | Q_t, y_{t-1})
\]
Following the results of~\citep{hershey2007approximating}, and by setting \( \alpha_i = \alpha_j = 1 \), we obtain:
\[
\mathbb{E}_{\hat{x}_t}[\log p(\hat{x}_t | Q_t, y_{t-1})] \propto - \sum^{P}_{i=0} \log \sum^{P}_{j=0} \exp \left\{ {\frac{||\hat{x}^{(i)}_t - \hat{x}^{(j)}_t||^2_2}{2\sigma^2_x}} \right\}
\]

\[
\mathbb{E}_{\hat{x}_t}[\log p(\hat{x}_t | Q_t, y_{t-1})] \propto -  \sum_{i,j} \log \left( \exp \left( \frac{\sum_{a \in Q_t} ([\hat{x}^{(i)}_t]_a - [\hat{x}^{(j)}_t]_a)^2}{2\sigma_x^2} \right) \right)
\]

Assuming $k$ is the set of potential measurement locations at time step t, and $Q_t = Q_{t-1} \cup q_t$, where $q_t \in k$.
\[
\mathbb{E}_{\hat{x}_t}[\log p(\hat{x}_t | Q_t, y_{t-1})] \propto - \sum_{i,j} \log \left( \exp \left( \frac{\sum_{q_t \in k} ([\hat{x}^{(i)}_t]_{q_t} - [\hat{x}^{(j)}_t]_{q_t})^2 + \sum_{r \in Q_{t-1}} ([\hat{x}^{(i)}_t]_r - [\hat{x}^{(j)}_t]_r)^2}{2\sigma_x^2} \right) \right)
\]

\[
\mathbb{E}_{\hat{x}_t}[\log p(\hat{x}_t | Q_t, y_{t-1})] \propto - \sum_{i,j} \log \left( \prod_{q_t \in k} \exp \left( \frac{([\hat{x}^{(i)}_t]_{q_t} - [\hat{x}^{(j)}_t]_{q_t})^2}{2\sigma_x^2} \right) \prod_{r \in Q_{t-1}} \exp \left( \frac{([\hat{x}^{(i)}_t]_r - [\hat{x}^{(j)}_t]_r)^2}{2\sigma_x^2} \right) \right)
\]

\[
\mathbb{E}_{\hat{x}_t}[\log p(\hat{x}_t | Q_t, y_{t-1})] \propto - \sum_{i,j} \left( \sum_{q_t \in k} \frac{([\hat{x}^{(i)}_t]_{q_t} - [\hat{x}^{(j)}_t]_{q_t})^2}{2\sigma_x^2} + \sum_{r \in Q_{t-1}} \frac{([\hat{x}^{(i)}_t]_r - [\hat{x}^{(j)}_t]_r)^2}{2\sigma_x^2} \right)
\]

We then compute the expected log-likelihood at a specified measurement location $q_t$, discarding all terms independent of $q_t$. This key observation allows us to simplify the expression as follows:

\[
\underbrace{\mathbb{E}_{\hat{x}_t}[\log p(\hat{x}_t | Q_t, y_{t-1})] \biggr\rvert_{q_t}}_\text{The expected log-likelihood at a measurement location $q_t$} \propto  \sum_{i,j} \left( -\frac{([\hat{x}^{(i)}_t]_{q_t} - [\hat{x}^{(j)}_t]_{q_t})^2}{2\sigma_x^2} \right)
\]

Equivalently, we can write the above expression as:

\[
\mathbb{E}_{\hat{x}_t}[\log p(\hat{x}_t | Q_t, y_{t-1})] \biggr\rvert_{q_t} \propto  \left( \underbrace{\sum_{i=0}^{P} \sum_{j=0}^{P} \exp \left\{ - \frac{([\hat{x}^{(i)}_t]_{q_t} - [\hat{x}^{(j)}_t]_{q_t})^2}{2\sigma_x^2} \right\}}_{\mathrm{likeli}^{\mathrm{score}}(q_t)} \right)
\]

By definition, the left-hand side of the above expression corresponds to $\mathrm{likeli}^{\mathrm{score}}(q_t)$.
\end{proof}

\section{Empirical Analysis of $h$ model}
\subsection{Doob's $h$-transform as the Correction Factor}\label{app:10}
We demonstrate that the $h$-transform serves as a correction term for Tweedie's estimate. Specifically, the conditional Tweedie estimate can be expressed as:

\[
\mathbb{E}[x_0 \mid x_t, y] \approx \hat{x}_0(x_t, y)
= \frac{x_t - \sqrt{1-\bar{\alpha}_t} \left( h^{\zeta}_t(x_t, y) + s^{\theta^*}_t(x_t) \right)}{\sqrt{\bar{\alpha}_t}}
\]
\[
= \underbrace{\big(\frac{x_t}{\sqrt{\bar{\alpha}_t}} -\frac{\sqrt{1-\bar{\alpha}_t}}{\sqrt{\bar{\alpha}_t}} s^{\theta^*}_t\big)}_{\text{Unconditional Tweedie estimate}} - \underbrace{\frac{\sqrt{1-\bar{\alpha}_t}}{\sqrt{\bar{\alpha}_t}} h^{\zeta}_t(x_t, y)}_{\text{Correction Factor (i.e., $h$-transform)}} 
\]
In Equation (40), the first term represents the unconditional Tweedie estimate, illustrating that the $h$-transform can be viewed as a correction to the unconditional denoised prediction.

\subsection{Details of $h$-model Update Scheduler}\label{app:scheduler}
As highlighted in the main paper, the pessimistic updating of the h-model parameters plays a critical role in stabilizing the prior model’s adaptability dynamics. While a straightforward heuristic might suggest updating the h-model after a fixed number of observations, this approach only provides marginal improvements. A more effective scheduling strategy requires fewer updates during the early stages of discovery, allowing the model to gather ample data and avoid the risk of erroneous updates when the observations are still sparse. However, as the discovery process progresses and the model’s understanding of the search space strengthens, more frequent updates of the h-model become essential. This shift enables more effective exploitation of the environment, as the model has already gathered enough information, making the need for pessimistic updates unnecessary. Motivated by this observation, we propose the following $h$-model update scheduler:
\begin{equation}\label{eq:scheduler}
\Delta t_i = \frac{\mathcal{B}}{U} \cdot (1 - \frac{i}{U+1})^\gamma
\end{equation}
Here, $\mathcal{B}$ denotes the overall sampling budget, $U$ is the total number of $h$-model updates throughout the active discovery process, $i$ indicates the current sampling step, $\gamma$ governs the decay rate, and $\Delta t_i$ defines the interval between two successive updates of the $h$-model parameters at the $i$-th sampling step. Note that since $\gamma \geq 1$, the update frequency of the $h$-model naturally accelerates with increasing $i$, leading to more frequent updates during the later stages of the active discovery process.

\subsection{Effect of $h$-model Update Scheduler}
In this section, we evaluate the effectiveness of the $h$-model update scheduler by comparing two variants of EM-PTDM: one using a uniform update schedule and the other employing the adaptive scheduler defined in Equation~\ref{eq:scheduler}. For the uniform scheduler, the $h$-model is updated at fixed intervals—specifically, every 20 update steps. For the adaptive scheduler, we set $\gamma = 1$, $U = 30$, and an observation budget of $\mathcal{B} = \{ 200, 250\}$. The comparative results under this configuration are presented in Table~\ref{tab:h_adapt}. For this analysis, we consider active discovery of balls with a diffusion model trained on MNIST data as the prior model. Our empirical results show that \textbf{EM-PTDM with an adaptive $h$-model update scheduler consistently outperforms its uniform counterpart across various measurement budgets. This improvement can be attributed to fewer updates in the early stages, allowing the model to collect more informative observations before updating and thus reducing the risk of premature or noisy updates.} As the discovery progresses and the model gains a stronger understanding of the search space, the increased update frequency of $h$-model in later stages proves beneficial for accelerating active target discovery.

\begin{table}[!h]
  \vspace{-10pt}
  \centering
  \caption{Effect of Adaptive $h$-model Update Scheduler.} 
  \label{tab:h_adapt}
  \begin{tabular}{p{4.95cm}p{2.28cm}p{2.28cm}}
    \toprule
    \multicolumn{3}{c}{Active Discovery of Balls with MNIST Digit Images as the Prior.}  \\
    \midrule
    \small{$h$-model update Schedule} & $\mathcal{B}=200$ & $\mathcal{B}=250$ \\ 
    \midrule
    Uniform & 0.6856 & 0.7875 \\ 
    \textbf{Adaptive} & \textbf{0.7364} & \textbf{0.8268} \\ 
    \bottomrule
  \end{tabular}
\end{table}
\section{Exploratory Nature of EM-PTDM}
In active target discovery under an uninformative prior, exploration isn't just helpful—it's essential. Especially in the early stages, when the permanent memory offers little insight into the target domain and the correction factor is large, the $h$-model must rapidly adapt. This demands smart, strategic exploration of the search space to collect informative observations, enabling the $h$-model to efficiently learn and calibrate the correction factor. To assess EM-PTDM's exploratory behavior, we tackle active target discovery of overhead objects using ground-level imagery as prior knowledge, evaluating across a wide range of observation budgets—from very sparse (200) to less sparse (350)—and benchmark against baseline methods. We present the results in the following Table~\ref{tab:pm_exp}. 

\begin{table}[!h]
  \centering
  \caption{Importance of Exploration}
  \label{tab:pm_exp}
  \begin{tabular}{p{1.95cm}p{1.28cm}p{1.28cm}p{1.28cm}p{1.28cm}}
    \toprule
    \multicolumn{5}{c}{ATD of Overhead Objects with ImageNet as Prior.} \\
    \midrule
    Method & $\mathcal{B}=200$  & $\mathcal{B}=250$ & $\mathcal{B}=300$ & $\mathcal{B}=350$ \\
    \midrule
    DiffATD & 0.3873 & 0.5143 & 0.6391 & 0.7348 \\
    GA & 0.3479 & 0.4784 & 0.5659 & 0.6562   \\
    \hline 
    \textbf{\emph{EM-PTDM}} & \textbf{0.4127} & \textbf{0.5620} & \textbf{0.7013} & \textbf{0.8256}  \\ 
    \bottomrule
  \end{tabular}
\end{table}
We observe that EM-PTDM’s performance improvement over baselines grows with the observation budget. When the budget is low, the performance gap is narrow, reflecting limited opportunity for exploration. As the budget increases, EM-PTDM leverages richer exploration to adapt its $h$-model, leading to significantly more effective target discovery.

\section{Training and Inference Pseudocode}\label{app:pseudo}
Below we present the training and inference pseudocode.
\begin{algorithm}
\caption{\textsc{EM-PTDM Sampling Strategy (At the $t$-th Observation Step)}}
\label{alg:h-transform-ddim-sampling}
\textbf{Require:} Current State of Transient Memory: Trained $h$-transform $h^{\zeta}_t(x, \hat{x}_0, y)$ with parameters $\zeta$. \\
\textbf{Require:} Permanent Memory as Unconditionally trained noise predictor $s^{\theta^*}_t(x_t)$ \\
\textbf{Require:} Noise schedule $\beta_t = \beta(t)$, $\bar{\alpha}_t = \bar{\alpha}(t)$ \\
\textbf{Require:} Sampling schedule $\sigma_t = \sigma(t)$ \\
\textbf{Require:} Observation $y$, Posterior samples list $ps=$ [], Success = $R=0$. 
\begin{algorithmic}[1]
\State $x_T \sim P_T = \mathcal{N}(0, 1)$ \Comment{Sample a starting point}
\For{$i=$ $P$ to $1$}
  \State $\hat{x}^{i} = 0$
  \For{$t$ in $(T, T-1, \ldots, 1)$}
    \State $\hat{\epsilon}_{\theta} \gets s^{\theta^*}_t(x_t)$ \Comment{Predict unconditional noise}
    \State $\hat{x}_0 \gets \dfrac{x_t - \sqrt{1 - \bar{\alpha}_t} \hat{\epsilon}_{\theta}}{\sqrt{\bar{\alpha}_t}}$
    \State $\hat{\epsilon}_{\zeta} \gets h^{\zeta}_t(x_t, \hat{x}_0, y)$ \Comment{Predict correction noise via $h$-transform}
    \State $\hat{\epsilon} \gets \hat{\epsilon}_{\theta} + \hat{\epsilon}_{\zeta}$ \Comment{Estimate posterior noise}
    \If{$t > 1$}
        \State Sample $\epsilon_t \sim \mathcal{P}_{\text{noise}}$
    \Else
        \State $\epsilon_t \gets 0$
    \EndIf
    \State $x_{t-1} \gets \sqrt{\bar{\alpha}_{t-1}} \left( \dfrac{x_t - \sqrt{1 - \bar{\alpha}_t} \hat{\epsilon}}{\sqrt{\bar{\alpha}_t}} \right) + \sqrt{1 - \bar{\alpha}_{t-1} - \sigma_t^2} \hat{\epsilon} + \sigma_t \epsilon_t$
    \If{$t = 1$}
       \State $\hat{x}^{i} = x_{t-1}$.
    \EndIf
  \EndFor
  \State $ps$.append($\hat{x}^{i}$)
\EndFor
  \State Utilize Posterior samples in $ps$ to compute $\mathrm{expl}^{\mathrm{score}}(q_{t})$ and $\mathrm{exploit}^{\mathrm{score}}(q_t)$ using Eqn.~\ref{eq:exp-ecore} and~\ref{eq:exploit-score} respectively for each $q_t \in k$.
        \State Compute $\mathrm{score}(q_t)$ using Eqn.~\ref{eq:final_score} for each $q_t \in k$ and sample a location $q_t$ with the highest $\mathrm{score}$.
        \State $\mathcal{B} \gets \mathcal{B} - 1$, $\{ k \} \gets \{ k \} \setminus q_t$
        \State Update: $Q_{t} \gets Q_{t-1} \cup q_{t}$, $y_t \gets y_{t - 1} \cup [x]_{q_t}.$ 
        \State Update: $\mathcal{D}_{t} \gets \mathcal{D}_{t-1} \cup \{[x]_{q_{t}}, y^{(q_{t})}\}, R$ += $y^{(q_t)} $
        \State Train $r_{\eta}$ with updated $\mathcal{D}_{t}$ and optimize $\eta$ with Cross-Entropy loss.
\State \Return $R$
\end{algorithmic}
\end{algorithm}
\begin{algorithm}
\caption{\textsc{h-Transform Fine-Tuning (At the $t$-th Observation Step)}}\label{alg:htransform}
\begin{algorithmic}[1]
\Require Posterior Samples drawn from $q_{\phi_{t-1}}(x \mid y_{t-1})$ 
\Require Noise schedule $\beta_t = \beta(t)$, $\bar{\alpha}_t = \bar{\alpha}(t)$ 
\Require Permanent Memory (i.e. Pre-Trained Noise predictor function) $s^{\theta^*}_t(x)$ with parameters $\theta^*$.
\Require Current state of Transient Memory (i.e. $h$-transform) $h^{\zeta}_t(x, \hat{x}_0, y)$ with parameters $\zeta$.
\Repeat
    \State $x_0 \sim P_0 = q_{\phi_{t-1}}(x \mid y_{t-1})$
    \State $t \sim \text{Uniform}(\{1, \ldots, T\})$
    \State $\varepsilon_t \sim \mathcal{N}(0, I)$ \Comment{Sample noise}
    \State $x_t \leftarrow \sqrt{\bar{\alpha}_t}x_0 + \sqrt{1-\bar{\alpha}_t}\varepsilon_t$
    \State $\hat{\varepsilon}_\theta \leftarrow s^{\theta^*}_t(x_t)$ \Comment{Estimate noise with pretrained model}
    \State $\hat{x}_0 \leftarrow \frac{x_t - \sqrt{1-\bar{\alpha}_t}\hat{\varepsilon}_\theta}{\sqrt{\bar{\alpha}_t}}$
    \State $\hat{\varepsilon}_\zeta \leftarrow h^{\zeta}_t(x_t, \hat{x}_0, y)$ \Comment{Estimate correction via h-transform}
    \State Take gradient descent step w.r.t. $\zeta$ on \\
    \>\>\>\>\>\>\>\>$\nabla_\zeta \mathcal{L}(\varepsilon_t, \hat{\varepsilon}_\theta + \hat{\varepsilon}_\zeta)$
    \Comment{$\mathcal{L}$ is defined in Equation~\ref{eq:sm}}
\Until{convergence or maximum epochs reached}\\
\Return Updated $h-$model Parameter $\zeta$.
\end{algorithmic}
\end{algorithm}

\section{Efficacy of $h$-model in Estimating The Search Space with only Few Observations}
\begin{figure}[htbp]
  \centering
  \begin{subfigure}[b]{0.45\textwidth}
    \includegraphics[width=\textwidth]{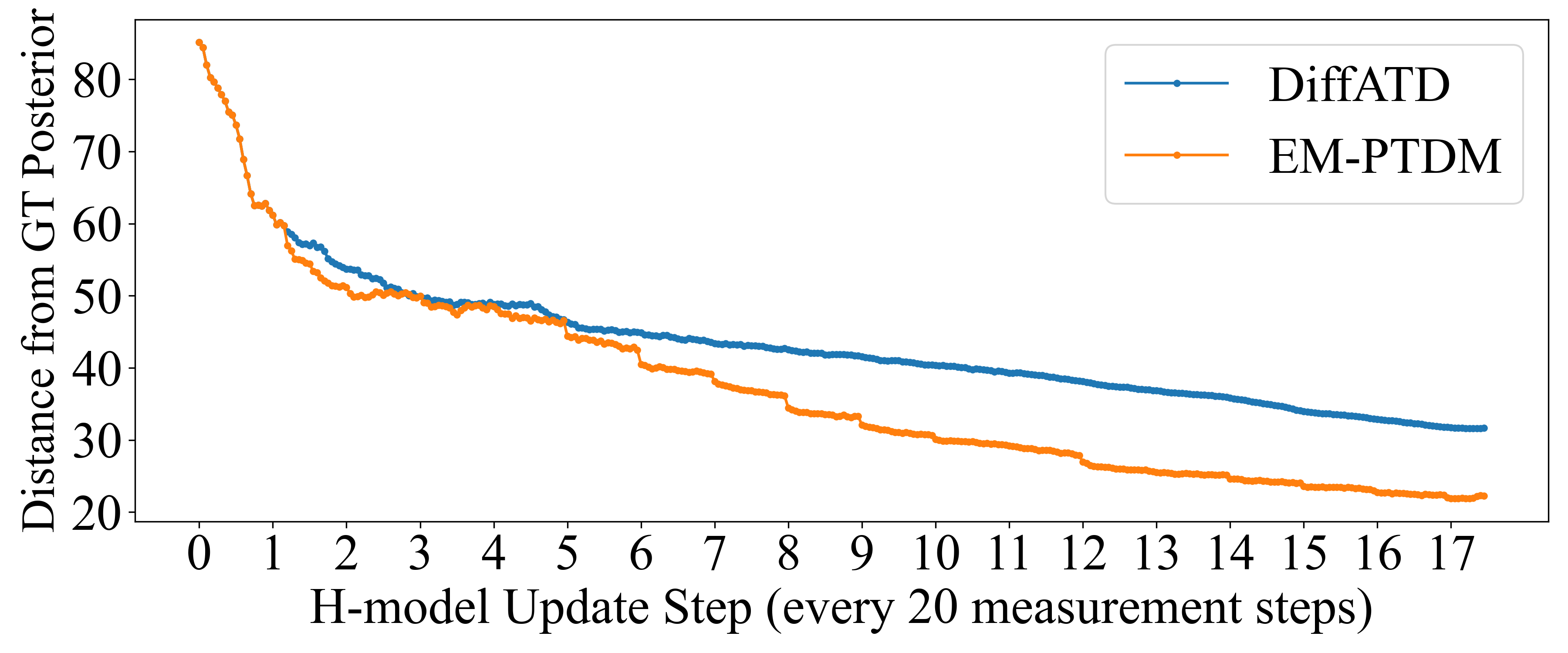}
    \caption{L2 Distance between predicted and ground-truth posterior.}
    \label{fig:left}
  \end{subfigure}
  \hfill
  \begin{subfigure}[b]{0.45\textwidth}
    \includegraphics[width=\textwidth]{figures/update_memory_5.png}
    \caption{L2 Distance between predicted and ground-truth targets.}
    \label{fig:right}
  \end{subfigure}
  \caption{Quantitative analysis of $h$-model's adaptability in Active Target Discovery.}
  \label{fig:quant_analysis}
\end{figure}

We analyze the adaptability of the $h$-model using quantitative visualizations in the context of an active target discovery of overhead objects, where ground-level ImageNet images serve as the prior. Specifically, we assess the $h$-model’s role by computing L2-based semantic similarity between predicted and ground-truth targets (right~\ref{fig:right}), and between predicted and ground-truth posteriors (left~\ref{fig:left}). As depicted in Figure~\ref{fig:quant_analysis}, the inclusion of the $h$-model leads to significantly faster convergence to the true posterior and the true targets, thus enabling more informed target discovery. \textbf{The rapid drop in dissimilarity compared to using permanent memory alone highlights the $h$-model’s ability to quickly correct the prior with minimal observations.}

\section{Analyzing the Role of Permanent and Transient Memory for Enhancing In-Domain Target Discovery }
Since we've seen that inside the EM-PTDM framework, \textbf{updating permanent memory with posteriors from prior in-domain ATD tasks boosts performance, it raises a key question: do we still need the $h$-model?} To investigate, we turn to DiffATD—a state-of-the-art baseline that uses only permanent memory. To update the permanent memory of DiffATD, we apply the same continual memory update strategy as in EM-PTDM, incorporating accumulated posteriors after each task, to see how far performance can go without the $h$-model. For this comparison, we examine active target discovery of overhead objects using ground-level ImageNet images as prior knowledge. While updating only the permanent memory leads to improved discovery rates compared to DiffATD with fixed permanent memory across different observation budgets, a clear and consistent performance gap remains when compared to EM-PTDM, particularly when EM-PTDM updates its permanent memory after each in-domain ATD task. \textbf{These results, summarized in Table~\ref{tab:h_pm_upd}, highlight the added value of the $h$-model in driving more effective exploration and adaptation irrespective of whether permanent memory is being updated or not.} 

\begin{table}[!h]
  \vspace{-10pt} 
  \centering
  \caption{Importance of $h$-model with or without Permanent Memory (\textbf{PM}) Update}
  \label{tab:h_pm_upd}
  \begin{tabular}{p{3.95cm}p{1.28cm}p{1.28cm}p{1.28cm}}
    \toprule
    \multicolumn{4}{c}{ATD of Overhead Objects with ImageNet as Prior.} \\
    \midrule
    Method & $\mathcal{B}=250$ & $\mathcal{B}=300$ & $\mathcal{B}=350$ \\
    \midrule
    DiffATD & 0.5143 & 0.6391 & 0.7348 \\ 
    DiffATD w/ \textbf{PM} Update & 0.5294 & 0.6623 & 0.7589 \\
    EM-PTDM & 0.5620 & 0.7013 & 0.8256 \\
    \textbf{EM-PTDM w/ PM Update} & \textbf{0.5859} & \textbf{0.7194} &\textbf{0.8461} \\
    \bottomrule
  \end{tabular}
  \vspace{-10pt} 
\end{table}

\section{Effect of $\kappa(\mathcal{B}$) }\label{par:kappa}
We conduct experiments to assess the impact of $\kappa(\mathcal{B})$ on EM-PTDM's active discovery performance. Specifically, we investigate how amplifying the exploration weight,
\begin{table}[!h]
  \vspace{-10pt} 
  \centering
  \caption{Effect of $\kappa(\mathcal{B})$}
  \label{tab:balance}
  \begin{tabular}{p{2.7cm}p{1.4cm}p{1.35cm}p{1.35cm}p{1.35cm}}
    \toprule
    \multicolumn{5}{c}{Performance across varying $\alpha$ with $\mathcal{B}=250$} \\
    \midrule
    Target  & Prior & $\alpha=0.2$ & $\alpha=1.0$ & $\alpha=5.0$ \\
    \midrule
    Balls & MNIST & 0.7416 & 0.7875 & \textbf{0.9272} \\
    \bottomrule
  \end{tabular}
  \vspace{-10pt} 
\end{table}
 by setting $\kappa(\mathcal{B}) = \max{\{0, \kappa(\alpha \cdot \mathcal{B})\}}$ with $\alpha > 1$, and enhancing the exploitation weight by setting $\alpha < 1$, influence the overall effectiveness of the approach. 
We report results for $\alpha \in \{0.2, 1, 5\}$ in two settings: using overhead objects as targets with ground-level images as the prior (first row), and discovering target balls using MNIST digit images as the prior (second row). The results are summarized in Table~\ref{tab:balance}. The best performance is achieved with $\alpha = 5$, and \textbf{the results suggest that 
higher values of $\alpha$ boosts performance, reinforcing the fact that exploration is key to success in active target discovery under an uninformative prior.}

\section{EM-PTDM's Capability of Discovering Isolated Targets within Observation Budget}

\begin{figure}[!h]
\includegraphics[width=\linewidth]{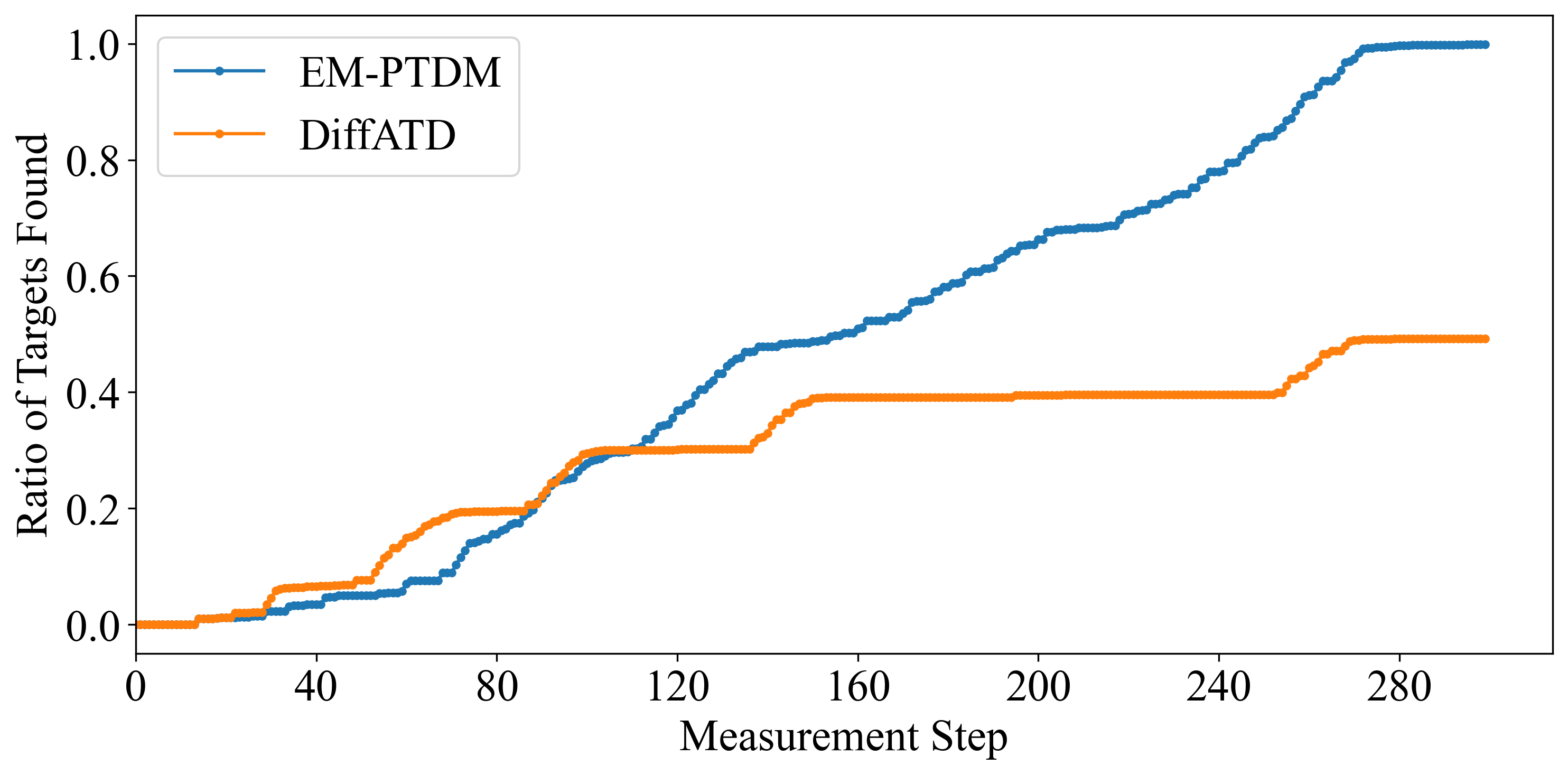}
\caption{Comparison of the Discovery Process: EM-PTDM vs. DiffATD. In this experiment, we evaluate the active discovery of 10 target balls using a diffusion model trained on MNIST images as the prior. As the search budget increases, EM-PTDM consistently uncovers more disjoint target balls, thanks to its inherently exploratory behavior. In contrast, DiffATD struggles to discover as many targets as it lacks the same efficiency as EM-PTDM in exploring the search space.}
\label{fig:ptdm_discover_curve}
\end{figure}

\begin{figure}[!h]
\includegraphics[width=\linewidth]{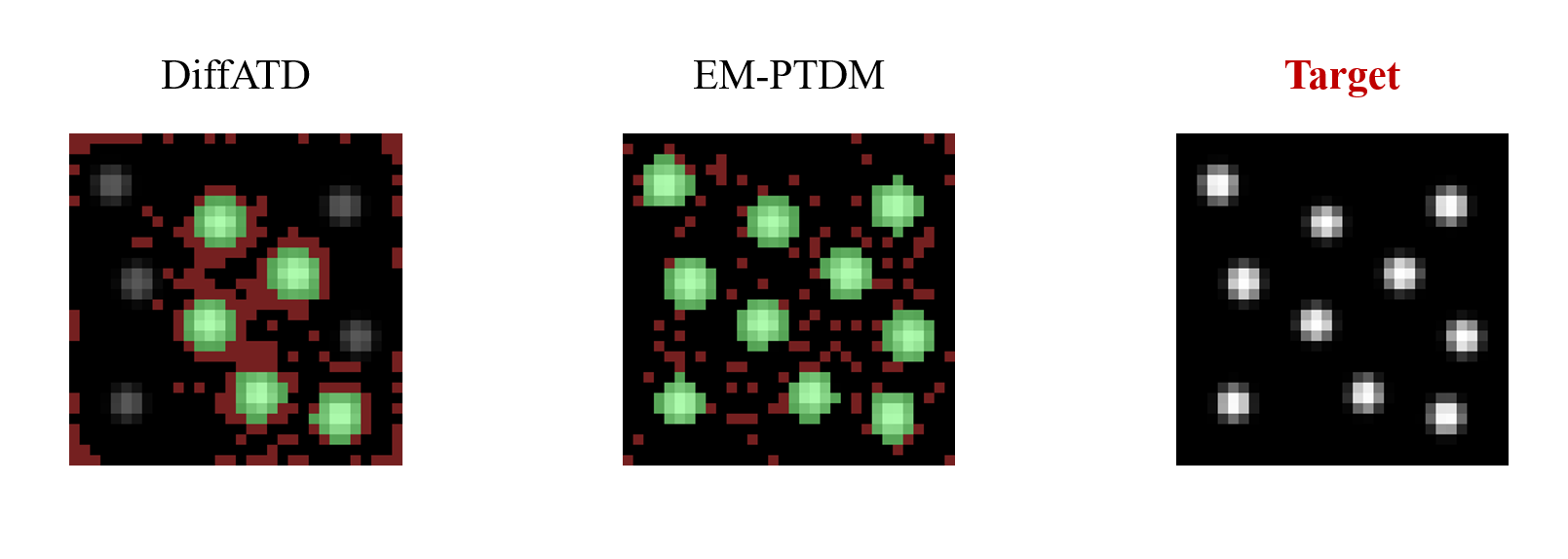}
\caption{Comparison between EM-PTDM and DiffATD's Discovery. \textbf{Green} patches correspond to successful target discovery, and \textbf{Red} Patches correspond to unsuccessful observations. In this example, the task is to discover the balls with MNIST images as the prior. The exploratory behavior of EM-PTDM, in contrast to DiffATD, is evident in the visualization.}
\label{fig:comparison_discovery_exp}
\end{figure}

To evaluate EM-PTDM’s ability to uncover disjoint target regions within a limited sampling budget, we design a series of controlled toy experiments. In each task, the goal is to discover a varying number of balls—positioned differently and with different radii—using a diffusion model pretrained on MNIST digits as the permanent memory. We systematically increase task difficulty by varying the number of target balls from 5 to 10. Notably, tasks with more targets demand effective exploration of the search space to successfully locate all disjoint regions within the budget constraints. 
We present comparative visualizations of the exploration behavior of EM-PTDM and DiffATD with an active discovery task involving uncovering 10 target balls with MNIST Images as the prior, 
shown in Fig.~\ref{fig:comparison_discovery_exp}. As the number of disjoint targets increases, making the task more challenging and exploration-intensive, EM-PTDM consistently succeeds in discovering most, if not all, targets within the given budget. In contrast, the baseline (i.e., DiffATD) relying solely on permanent memory struggles as task complexity rises, as seen in Plot ~\ref{fig:ptdm_discover_curve}. \textbf{These visualizations clearly demonstrate EM-PTDM’s superior exploration capabilities, which are crucial for efficient target discovery in settings with multiple disjoint targets.}

\vspace{-5pt}
\section{Species Distribution Modelling as Active Target Discovery Problem}\label{app:sde}
We constructed our species distribution experiment using observation data of the chosen species from iNaturalist. Center points were randomly sampled within North America (latitude 25.6°N to 55.0°N, longitude 123.1°W to 75.0°W). Around each center, we defined a square region approximately 480 km × 480 km in size (roughly 5 degrees in both latitude and longitude).  Each retained region was discretized into a 64×64 grid, where the value of each cell represents the number of observed species. To simulate the querying process, each 2×2 block of grid cells was treated as a query. 
\section{ Additional Results on Active Discovery of Unknown Species from Known Species Distribution}
\begin{figure}[!h]
\includegraphics[width=\linewidth]{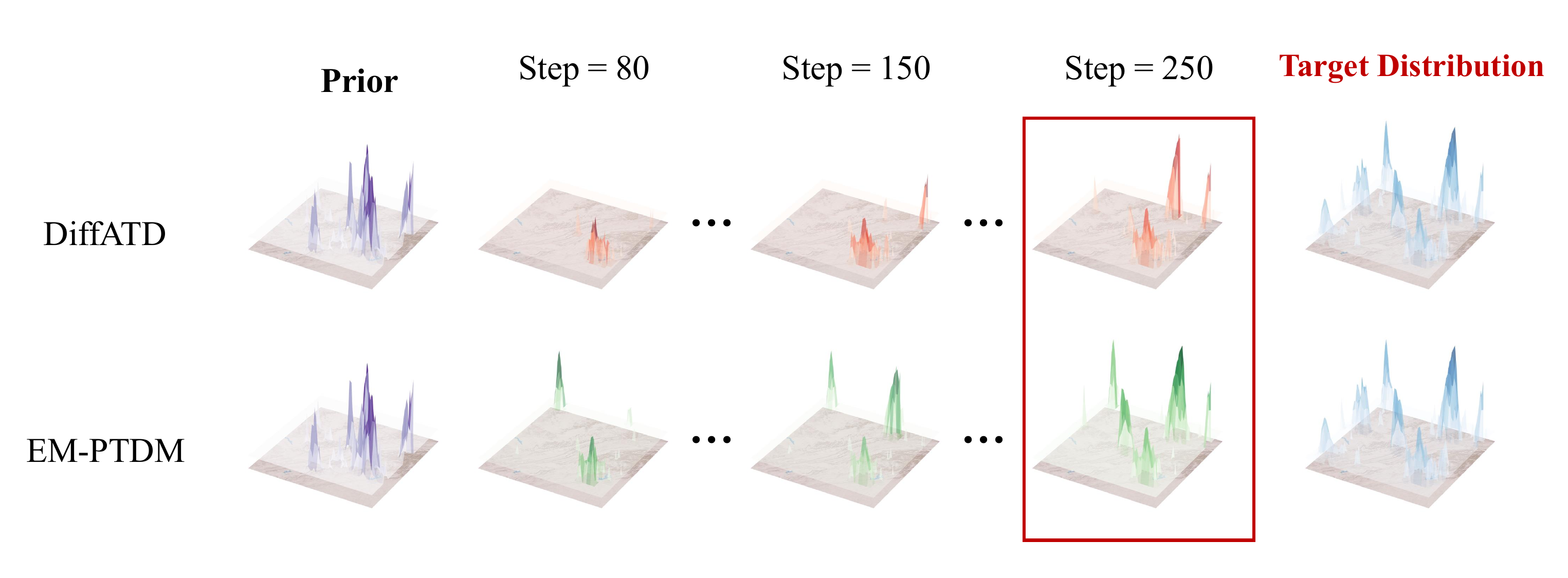}
\caption{Exploration Behavior of Different Approaches. We visualize the explored Regions at Different Active Target Discovery Phases. We consider the task of Active Discovery of the species Cedar Waxwing with a known distribution of the species Black-capped Chickadee. EM-PTDM discovers most target regions (i.e., more accurately discovers the existence of the species Cedar Waxwing).}
\label{fig:vis_species}
\end{figure}

In the main paper, we explored active discovery of Coccinella septempunctata (CS) using the known species distribution of Gladicosa and Gonioctena (GG) as the prior. This section extends our analysis to a different species from the iNaturalist dataset. Specifically, we evaluate EM-PTDM on a task that involves discovering Species Cedar Waxwing from the known distribution of Species Black-capped Chickadee. Figure~\ref{fig:vis_species} compares the exploration behavior of EM-PTDM and DiffATD at various stages of target discovery. As shown, EM-PTDM—starting from the same prior as DiffATD—progressively and efficiently adapts the prior toward the true target distribution with only a few task-specific observations. In contrast, DiffATD, which relies solely on static permanent memory, struggles to approximate the ground-truth distribution within the given observation budget.

\begin{table}[!h]
  \centering
  \captionof{table}{\emph{SR} Comparison with Baselines.}
  \begin{tabular}{p{2.89cm}p{2.70cm}p{2.70cm}p{2.70cm}}
    \toprule
    \multicolumn{4}{c}{Active Discovery of Cedar Waxwing Species with Species Black-capped Chickadee as Prior.} \\
    \midrule
    Method & $\mathcal{B}=150$ & $\mathcal{B}=200$ & $\mathcal{B}=250$ \\
    \midrule
    DiffATD & 0.2319 & 0.3309 & 0.4453 \\
    GA & 0.2232 & 0.3098 & 0.3116  \\
    \hline 
    \textbf{\emph{EM-PTDM}} & \textbf{0.3079} & \textbf{0.3854} & \textbf{0.6347}  \\ 
    \bottomrule
  \end{tabular}
  \label{tab:oth_spe}
\end{table}

We further compare the performance of EM-PTDM against baseline methods using Success Rate (SR) as the evaluation metric, with results summarized in Table~\ref{tab:oth_spe}. The task involves actively discovering Species Cedar Waxwing from the known distribution of Species Black-capped Chickadee. Consistent with other experimental settings, EM-PTDM significantly outperforms all baselines across varying observation budgets. These results further highlight the effectiveness of EM-PTDM in tackling active target discovery under an uninformative prior.

\section{More Visualizations on Efficiency of $h$-model's Adaptability From Very Sparse Observations }
\begin{figure}[!h]
\includegraphics[width=\linewidth]{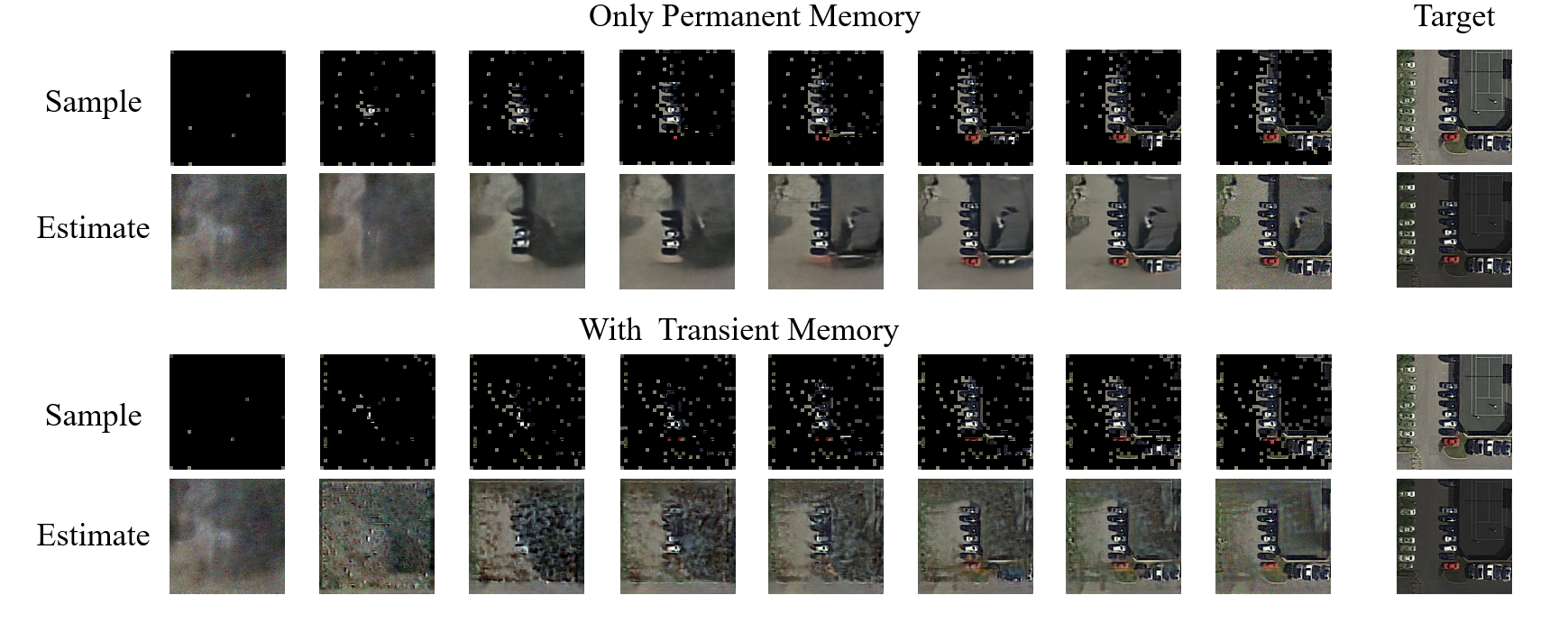}
\caption{$h$-model's Adaptability From Very Sparse Observations. Posterior Samples at Different Active Target Discovery Phases. Overhead Object Discovery (i.e., \textbf{Car}) with Ground Level images from ImageNet as the Prior. Observation Budget of 300.}
\label{fig:hvis_h_1}
\end{figure}

\begin{figure}[!h]
\includegraphics[width=\linewidth]{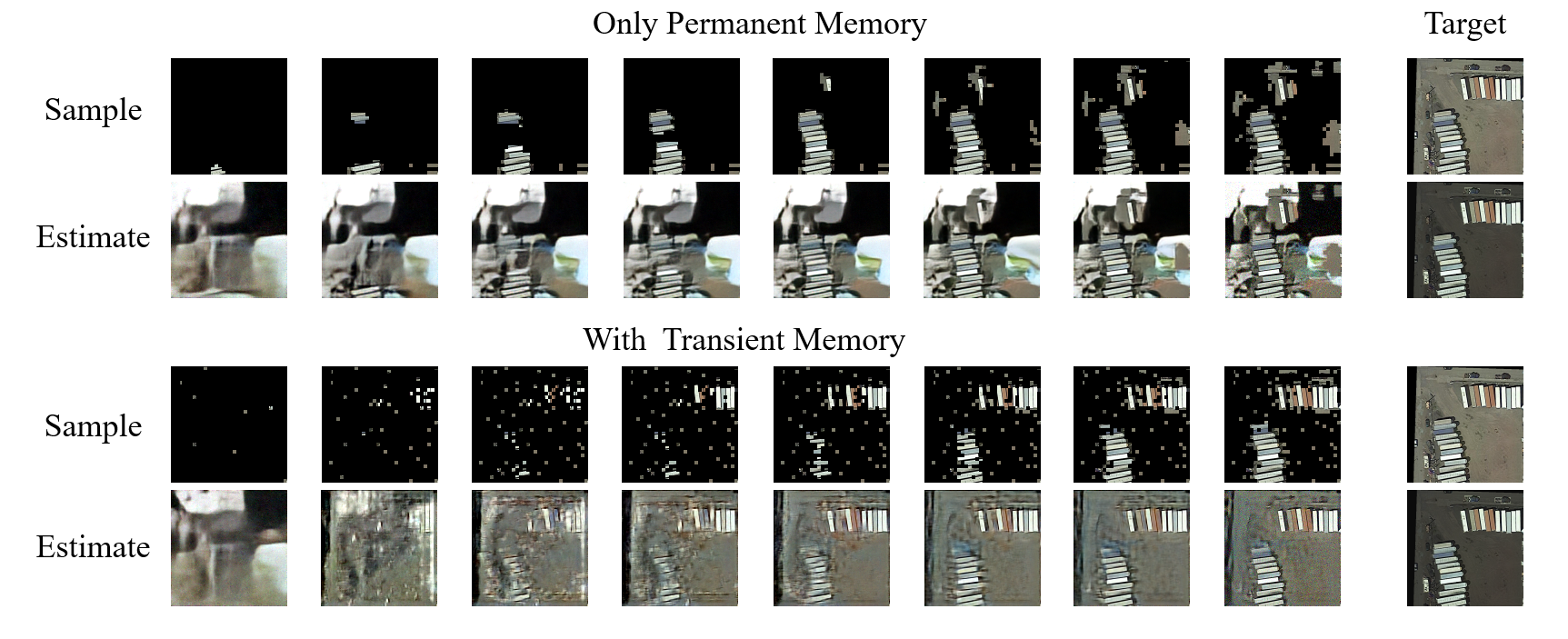}
\caption{$h$-model's Adaptability From Very Sparse Observations. Posterior Samples at Different Active Target Discovery Phases. Overhead Object Discovery (i.e., \textbf{Truck}) with Ground Level images from ImageNet as the Prior. Observation Budget of 300.}
\label{fig:hvis_h_2}
\end{figure}

\begin{figure}[!h]
\includegraphics[width=\linewidth]{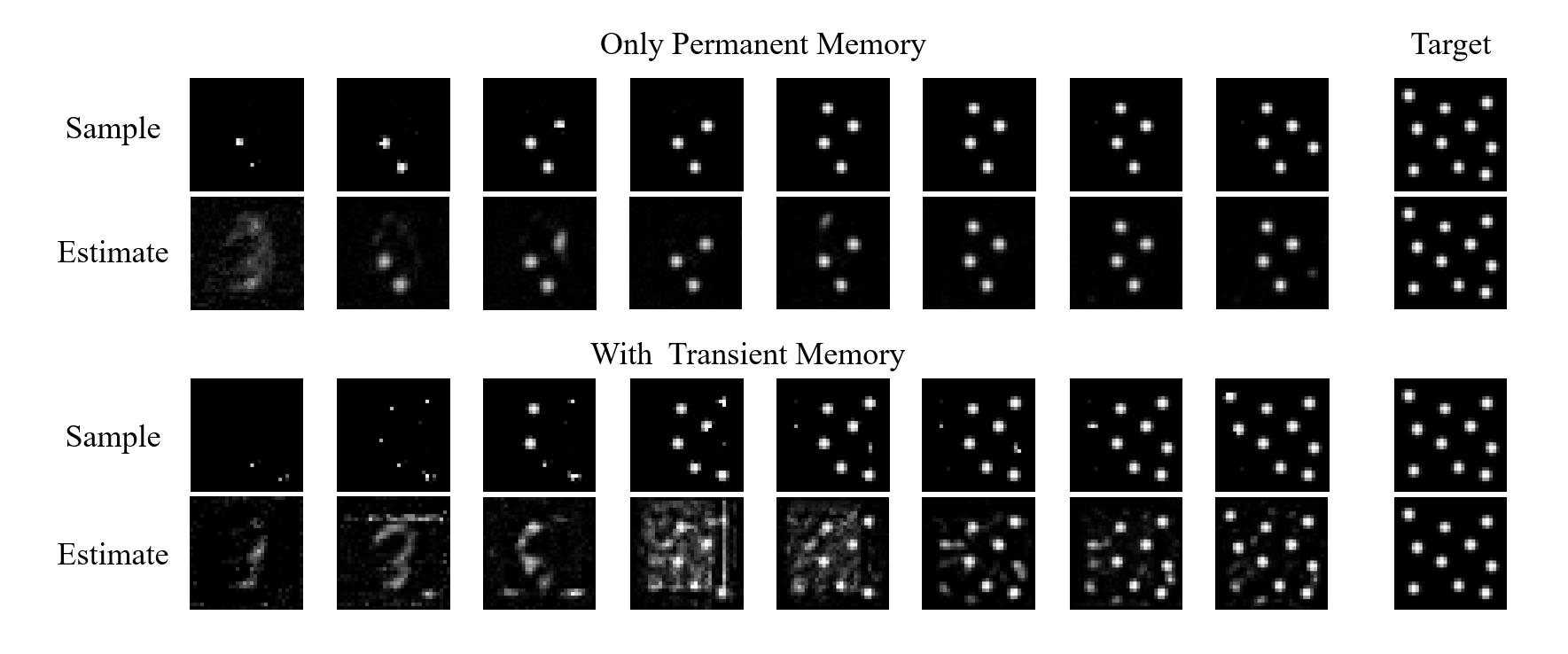}
\caption{$h$-model's Adaptability From Very Sparse Observations. Posterior Samples at Different Active Target Discovery Phases. Uncovering Disjoint Balls with MNIST digit images as the Prior.}
\label{fig:hvis_h_3}
\vspace{-6pt}
\end{figure}

In this section, we provide additional visualizations of posterior samples generated by EM-PTDM and DiffATD across different stages of active target discovery. As shown in Figures (~\ref{fig:hvis_h_1},~\ref{fig:hvis_h_2},~\ref{fig:hvis_h_3}), EM-PTDM produces samples that are more semantically aligned with the ground-truth posterior compared to DiffATD. Notably, even with sparse observations, EM-PTDM effectively simulates the search space, enabling more informed exploration and leading to improved target discovery under an uninformative prior.
\newpage

\section{More Visualizations of the Exploration Behavior of EM-PTDM at Different Active Target Discovery Phases}

\begin{figure}[!h]
\includegraphics[width=\linewidth]{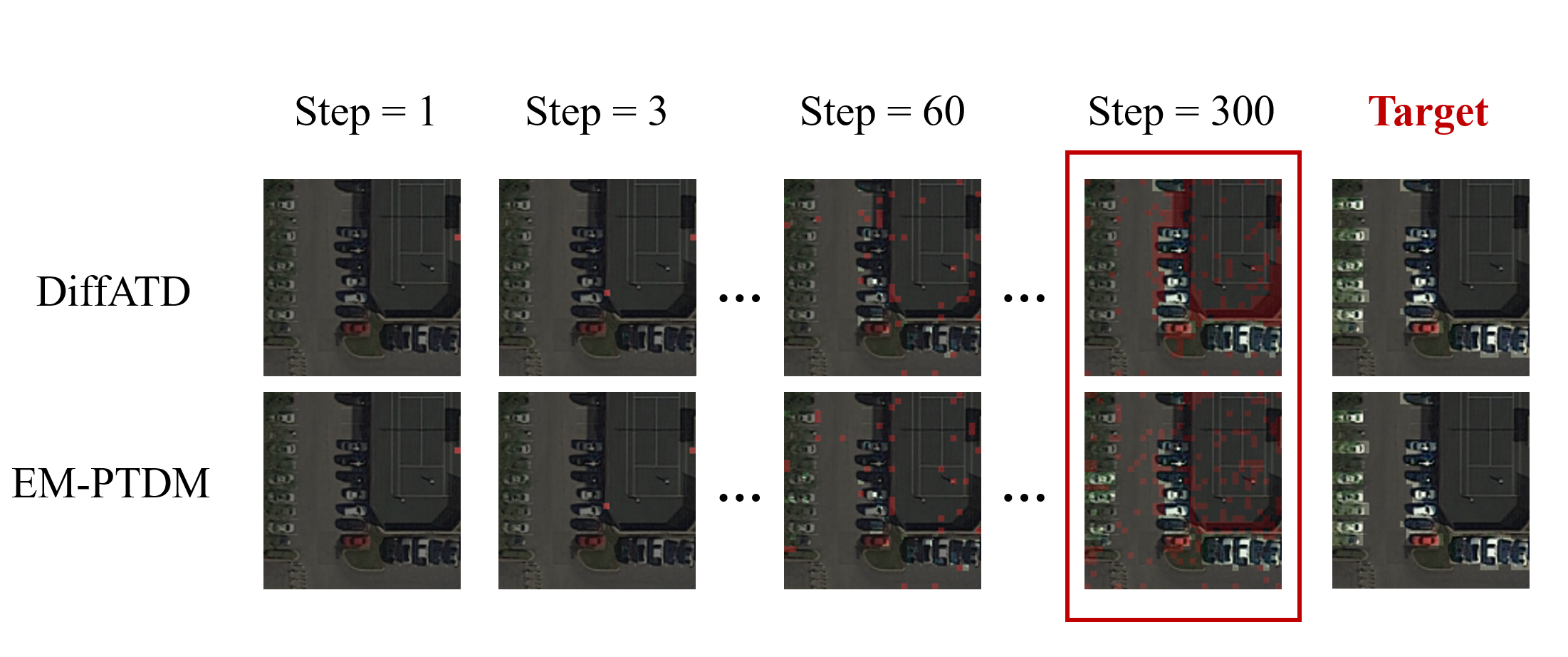}
\caption{Visualizing the regions explored by each method at various stages of the ATD process. In these visualizations, patches corresponding to \textbf{successful queries are unmasked}, while those resulting in \textbf{unsuccessful queries are highlighted in Red}. The task focuses on discovering overhead objects (\textbf{cars}), using ground-level images from ImageNet as the prior. The results demonstrate that EM-PTDM effectively identifies and explores most of the target regions containing \textbf{cars}.}
\label{fig:vis_exp_1}
\vspace{-6pt}
\end{figure}

\begin{figure}
\includegraphics[width=\linewidth]{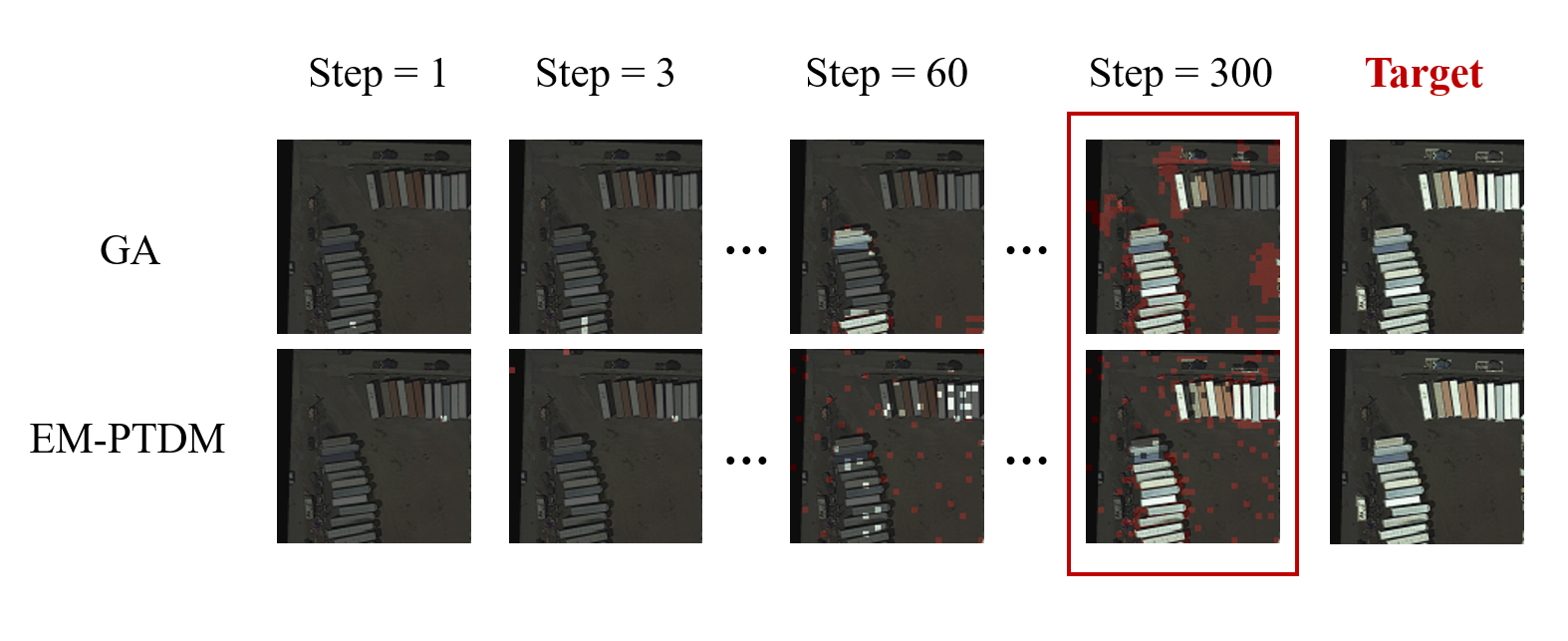}
\caption{Visualizing the regions explored by each method at various stages of the ATD process. In these visualizations, patches corresponding to \textbf{successful queries are unmasked}, while those resulting in \textbf{unsuccessful queries are highlighted in Red}. The task focuses on discovering overhead objects (\textbf{trucks}), using ground-level images from ImageNet as the prior. The results demonstrate that EM-PTDM effectively identifies and explores most of the target regions containing \textbf{trucks}.}
\vspace{-6pt}
\label{fig:vis_exp_2}
\end{figure}

\begin{figure}
\includegraphics[width=\linewidth]{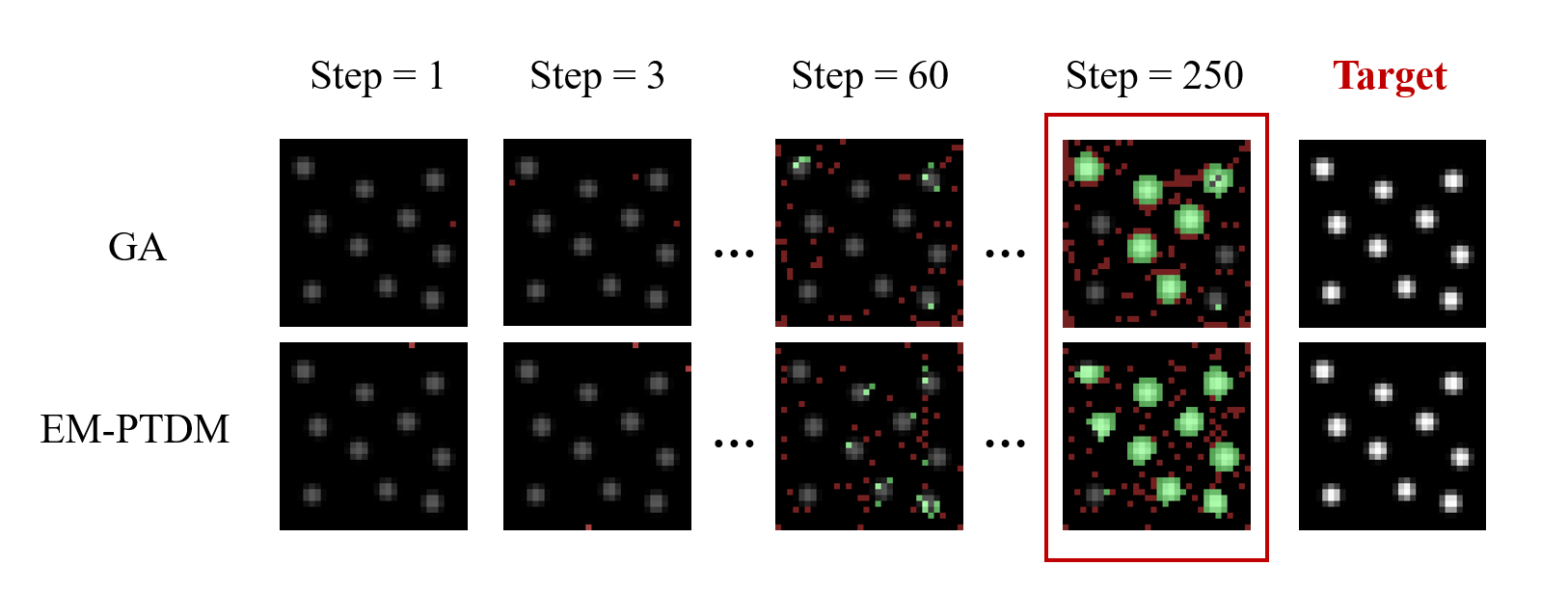}
\caption{\small{Visualizing the regions explored by each method at various stages of the ATD process. In these visualizations, patches corresponding to \textbf{successful queries are Green}, while those resulting in \textbf{unsuccessful queries are highlighted in Red}. The task focuses on uncovering Disjoint Balls from MNIST digit images as the Prior. EM-PTDM discovers most target regions (i.e., Disjoint Balls).}}
\label{fig:vis_exp_3}
\vspace{-6pt}
\end{figure}

In this section, we present additional exploration behavior of EM-PTDM at different active target discovery phases. We also provide a similar exploration behavior of the baseline approaches, including DifffATD and Greedy Adaptive, for the comparison. These visualizations are provided in Figures~\ref{fig:vis_exp_1},~\ref{fig:vis_exp_2},~\ref{fig:vis_exp_3}. These additional visualizations further reinforce the effectiveness of EM-PTDM in addressing active target discovery under an uninformative prior. 
\vspace{-6pt}
\vspace{-6pt}
\clearpage

\vspace{10pt}

\newpage
\section{Active Target Discovery of Balls Using MNIST Images as the Prior }
To enable a comprehensive analysis of EM-PTDM, we introduce a custom-designed dataset tailored for this task. It simulates active discovery scenarios involving an unknown number of balls with unknown locations and radii, with MNIST images as the prior. Success in this setting requires effective exploration of the search space to accurately localize the targets, capturing the core challenge of the problem. The details of the proposed dataset are provided below.

\subsection{Dataset Creation Procedure}
In this dataset, we generate each sample by randomly placing 5 to 10 identical balls within a 32×32 2D grid. The radius of all balls in a given sample is either 3 or 4 pixels, randomly selected per sample. All placements are performed uniformly at random, subject to the non-overlapping constraint and the boundary condition that each ball lies entirely within the 32×32 space.  

\subsection{SR Comparisons with Baseline Approaches}
As in previous settings, we quantitatively evaluate EM-PTDM and baseline methods using the Success Rate (SR) metric. In this experiment, the task involves actively discovering target balls using a diffusion model trained on the MNIST dataset as the prior. The results, summarized in Table~\ref{tab:SR_MNIST}, show a consistent trend: EM-PTDM significantly outperforms all baselines across different measurement budgets. This further reinforces the effectiveness of EM-PTDM in handling active target discovery under an uninformative prior.

\begin{table}[!h]
  \centering
  \captionof{table}{\emph{SR} Comparison with Baselines.}
  \vspace{-2pt}
  \begin{tabular}{p{2.89cm}p{1.70cm}p{1.70cm}p{1.70cm}}
    \toprule
    \multicolumn{4}{c}{Active Discovery of balls with MNIST Digit Images as Prior.} \\
    \midrule
    Method & $\mathcal{B}=150$ & $\mathcal{B}=200$ & $\mathcal{B}=250$ \\
    \midrule
    RS & 0.1458 & 0.1826 & 0.2187  \\
    DiffATD & 0.4362 & 0.4432 & 0.4929 \\
    GA & 0.3250 & 0.5170 & 0.6257  \\
    \hline 
    \textbf{\emph{EM-PTDM}} & \textbf{0.5561} & \textbf{0.6856} & \textbf{0.7875}  \\ 
    \bottomrule
  \end{tabular}
  \label{tab:SR_MNIST}
\end{table}

\subsection{Analyzing the Exploration Strategies of EM-PTDM and DiffATD Under Increasing Task Complexity}
\begin{figure}[!h]
\includegraphics[width=\linewidth]{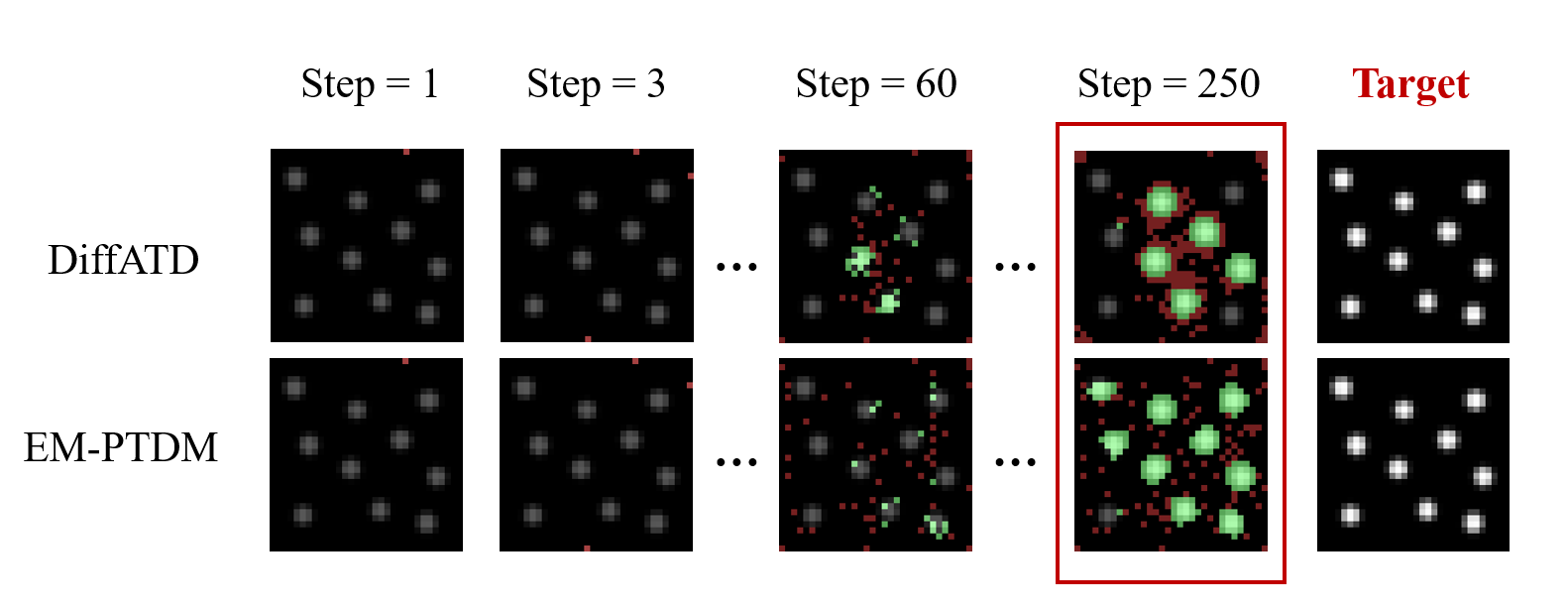}
\caption{Visualizing the regions explored by each method at various stages of the ATD process. In these visualizations, patches corresponding to \textbf{successful queries are highlighted in Green}, while those resulting in \textbf{unsuccessful queries are highlighted in Red}. The task focuses on uncovering Disjoint Balls from MNIST digit images as the Prior. The results demonstrate that EM-PTDM discovers most target regions (i.e., Disjoint Balls).}
\label{fig:vis_exp_ball}
\end{figure}

\begin{figure}[!h]
\includegraphics[width=\linewidth]{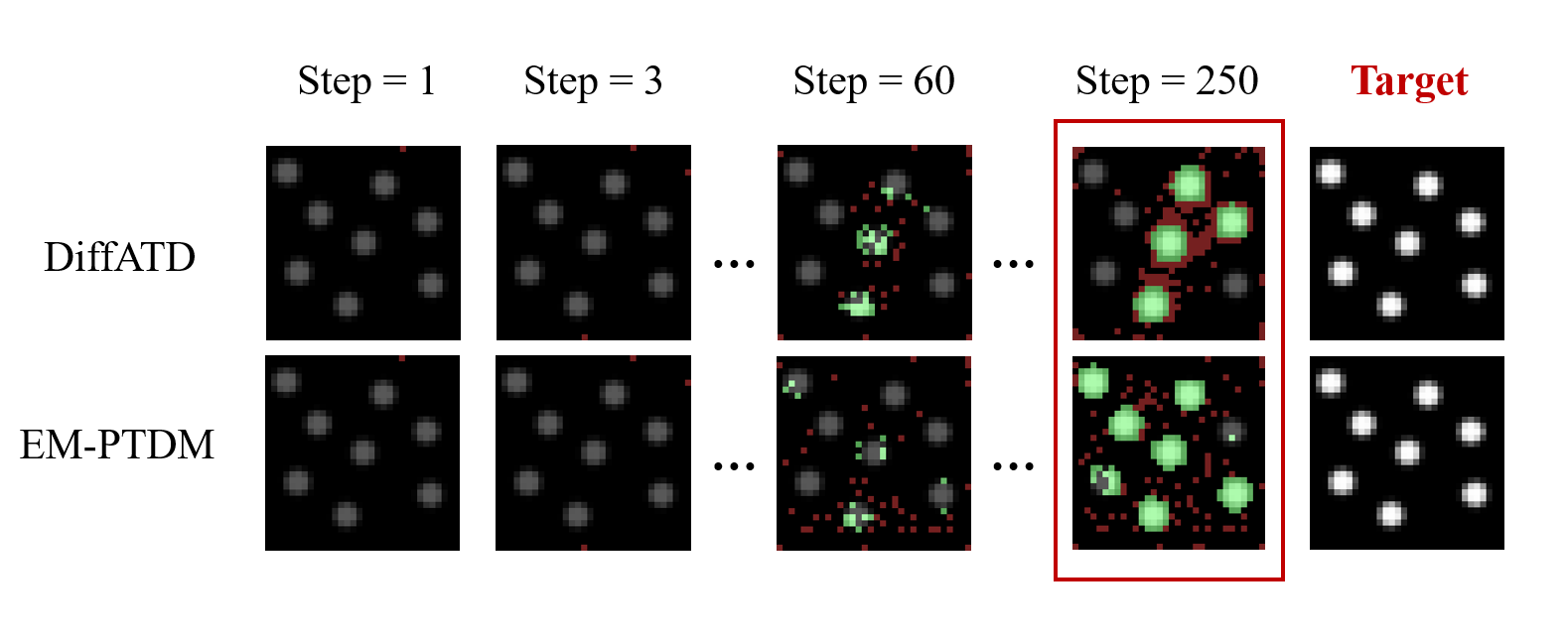}
\caption{Visualizing the regions explored by each method at various stages of the ATD process. In these visualizations, patches corresponding to \textbf{successful queries are highlighted in  Green}, while those resulting in \textbf{unsuccessful queries are highlighted in Red}. The task focuses on uncovering Disjoint Balls from MNIST digit images as the Prior. The results demonstrate that EM-PTDM discovers most target regions (i.e., Disjoint Balls).}
\label{fig:vis_exp_ball1}
\end{figure}

In this section, we provide additional visualizations highlighting the exploration behavior of EM-PTDM and DiffATD across different stages of the active target discovery task. Using the task of discovering target balls with MNIST images as the prior, the visualizations in Figures~\ref{fig:vis_exp_ball},~\ref{fig:vis_exp_ball1} clearly show that EM-PTDM consistently explores more effectively and identifies targets with higher accuracy, even under increasing task complexity while adhering to a strict budget and under an uninformative prior. These results further underscore the robustness and adaptability of EM-PTDM in challenging discovery scenarios, where efficient exploration of the search space is the key. \textbf{A striking emergent behavior is observed across both examples: in the early stages of the active discovery process, EM-PTDM engages in broader exploration of the search space compared to DiffATD. This initially results in fewer target discoveries (e.g., at step 60). However, this strategic exploration enables EM-PTDM to build a richer understanding of the environment, which it later exploits to surpass DiffATD, ultimately identifying a greater number of target regions before the observation budget is depleted (see step 250).}

\newpage

\section{Architecture, Training Details: $h-$model, Pretrained Diffusion Model, and the Reward Model; and Computing Resources}

\subsection{Details of $h$-model}
For the MNIST-to-Balls tasks, we employ 32-dimensional diffusion time-step embeddings and a single-block U-Net $h$-model with layer widths of [32, 64]. In the species discovery tasks, we also use 32-dimensional time-step embeddings, but with a single-block $h$-model featuring wider layers [32, 64, 128]. For the ImageNet-to-DOTA tasks, we increase the embedding dimensionality to 128, using a similar single-block $h$-model with widths [32, 64, 128] 

\subsection{Details of Reward Model}
Our proposed method, EM-PTDM, utilizes a parameterized reward model, 
$r_{\eta}$, to steer the exploitation process. To this end, we employ a neural network consisting of a series of convolutional and fully connected layers, with non-linear ReLU activations as the reward model ($r_{\eta}$). The reward model's goal is to predict a score ranging from 0 to 1, where a higher score indicates a higher likelihood that the measurement location corresponds to the target, based on its semantic features. 
Note that the size of the input semantic feature map for a given measurement location can vary depending on the downstream task. For instance, when working with DOTA, we use an $4 \times 4$ patch as the input feature size.
After each measurement step, we update the model parameters ($\eta$) using the binary cross-entropy loss. Additionally, the training dataset is updated with the newly observed data point, refining the model’s predictions over time. Naturally, as the search advances, the reward model refines its predictions, accurately identifying target-rich regions, which makes it progressively more dependable for informed decision-making. 
The reward model architecture 
consists of 1 convolutional layer with a $3 \times 3$ kernel, followed by 5 fully connected (FC) layers, each with its own weights and biases. The first FC layer maps an input of size $\frac{(input\>size)^2}{4}$ to an output of size 4 with weights and biases of size $[\frac{(input\>size)^2}{4}, 4]$ and $[4]$ respectively. The second FC layer transforms an input of dimension 4 to an output of size 32 with a 2-dimensional weight of size $[4, 32]$ and a bias of size $[32]$. The third FC layer maps 32 inputs to 16 outputs via a weight matrix of shape $[32, 16]$ and a bias vector of size $[16]$. The pre-final FC layer transforms inputs of size 16 to outputs of size 8  with $[16,8]$ weights, and a bias of shape $[8]$. The final FC layer produces an output of size 2, with weights of size $[8, 2]$ and a bias of size $[2]$, representing the target and non-target scores. The reward model uses the leaky ReLU activation function after each layer. We update the reward model parameters after each measurement step based on the binary cross-entropy loss. The reward model is trained incrementally for 3 epochs after each measurement step using the gathered supervised dataset resulting from sequential observation, with a learning rate of $0.01$.

\subsection{Details of Primary Memory as Pretrained Diffusion Model}
We use DDIM~\cite{song2020denoising} as the diffusion model across datasets. The diffusion models used in different experiments are based on widely adopted U-Net-style architecture. For the MNIST dataset, we use 32-dimensional diffusion time-step embeddings, with the diffusion model consisting of 2 residual blocks. We select the time-step embedding vector dimension to match the input feature size, ensuring the diffusion model can process it efficiently. The block widths are set to $[32, 64, 128]$, and training involves 30 diffusion steps. For DOTA, we use the input feature size of $[128, 128, 3]$, the architecture featuring 128-dimensional time-step embeddings and a diffusion model with 2 residual blocks of width $[64, 128, 256, 256, 512]$. 
Finally, all experiments are implemented in Tensorflow and conducted on NVIDIA A100 40G GPUs. Our training and inference code will be made public.

\section{Statistical Significance Results of EM-PTDM}

In order to strengthen our claim on EM-PTDM's superiority over the baseline methods, we have included the statistical significance results with different active target discovery settings, and present the results in Tables~\ref{tab:sg_dota},~\ref{tab:sg_species}. These results are based on 5 independent trials and further strengthen our empirical findings, reinforcing the stability and effectiveness of EM-PTDM in tackling active target discovery under an uninformative prior across diverse domains.
\begin{table}[H]
    \centering
    \caption{Statistical Significance Results for Unknown Overhead Object Discovery.}
    \begin{tabular}{p{3.5cm}p{2.5cm}p{2.5cm}p{2.5cm}}
        \toprule
        \multicolumn{4}{c}{Active Discovery of Overhead Objects with Ground Level ImageNet Images as the Prior.} \\
        \midrule
        Method & $\mathcal{B}=250$ & $\mathcal{B}=300$ & $\mathcal{B}=350$ \\
        \midrule
        RS & 0.2325 $\pm$ 0.0190 & 0.2852 $\pm$ 0.0137 & 0.3207 $\pm$ 0.0168  \\
        DiffATD & 0.5143 $\pm$ 0.0067 & 0.6391 $\pm$ 0.0102 & 0.7348 $\pm$ 0.0041 \\
        GA & 0.4784 $\pm$ 0.0122 & 0.5659 $\pm$ 0.0096 & 0.6562 $\pm$ 0.0054   \\
        \hline 
        \textbf{\emph{EM-PTDM}} & \textbf{0.5620} $\pm$ \textbf{0.0073} & \textbf{0.7013} $\pm$ \textbf{0.0038} & \textbf{0.8256} $\pm$ \textbf{0.0093}  \\ 
        \bottomrule
        \end{tabular}
        \label{tab:sg_dota}
\end{table}

\begin{table}[H]
    \centering
    \caption{Statistical Significance Results for Unknown Species Discovery Task.}
    \begin{tabular}{p{3.5cm}p{2.5cm}p{2.5cm}p{2.5cm}}
        \toprule
        \multicolumn{4}{c}{Active Discovery of Species CS with Species GG as the Prior.} \\
        \midrule
        Method & $\mathcal{B}=150$ & $\mathcal{B}=200$ & $\mathcal{B}=250$ \\
        \midrule
        RS & 0.1624 $\pm$ 0.0133 & 0.2327 $\pm$ 0.0201 & 0.2775 $\pm$ 0.0154  \\
        DiffATD & 0.3420 $\pm$ 0.0115 & 0.4365 $\pm$ 0.0057 & 0.4808 $\pm$ 0.0063 \\
        GA & 0.4061 $\pm$ 0.0047 & 0.5067 $\pm$  0.0079 & 0.5567 $\pm$ 0.0085  \\
        \hline 
       \textbf{\emph{EM-PTDM}} & \textbf{0.4983} $\pm$ \textbf{0.0060} &  \textbf{0.6495} $\pm$ \textbf{0.0108} & \textbf{0.6989} $\pm$ \textbf{0.0056}  \\ 
      \bottomrule
      \end{tabular}
      \label{tab:sg_species}
\end{table}

\section{Impact of Weak Permanent Memory on Active Target Discovery Performance}
When an extremely weak prior is used as the permanent memory, according to Equation (6), adapting to a new domain essentially reduces to learning the task from scratch through the transient memory alone. In this case, as indicated by Equation (7), the $h$-model no longer serves as a corrective mechanism; instead, under a partially observable environment, the entire responsibility for modeling the posterior shift falls on this lightweight module. However, this is beyond the capacity of such a lightweight module by design. In order to validate our hypothesis, we perform additional experiments under a remote sensing scenario, where we deliberately replaced the permanent memory with a diffusion model that could only output noise without any meaningful semantic structure. On top of this setup, we implemented EM-PTDM and observed the following phenomenon: even $h$-model was updated under partial observations, its limited capacity made it insufficient to compensate for the lack of a meaningful prior. As a result, all of the unexplored regions of the posterior estimation remained dominated by the weak prior—essentially resembling noise—leading to poor global environment estimation. We present our findings in the following Table~\ref{tab:weak_memory_analysis}. This supports the claim that \textbf{an extremely weak or non-semantic prior forces the transient memory to handle an unrealistically large modeling responsibility, which goes beyond its design capability.}

\begin{table}[h!]
\centering
\caption{Analysis of Weak Permanent Memory: Performance Comparison on DOTA}
\vspace{0.5em}
\begin{tabular}{lcc}
\toprule
\textbf{Method} & $\mathcal{B} = 250$ & $\mathcal{B} = 300$ \\ 
\midrule
EM-PTDM (\textbf{Random Noise} as Permanent Memory) & 0.3117 & 0.3465 \\ 
EM-PTDM (\textbf{ImageNet} as Permanent Memory) & \textbf{0.5620} & \textbf{0.7013} \\ 
\bottomrule
\end{tabular}
\label{tab:weak_memory_analysis}
\end{table}

\section{ More Details on Computational Cost across Search Space}
We have conducted a detailed evaluation of sampling time and computational requirements of EM-PTDM across various search space sizes. We present the results in the following table. Our results (as reported in~\ref{tab: COMP}) show that \textbf{EM-PTDM remains efficient even as the search space scales, with sampling time per observation step ranging from 0.83 to 1.87 seconds, which is well within practical limits for most downstream applications.} This further reinforces EM-PTDM’s scalability and real-world applicability.

\begin{table}[H]
    \centering
    \footnotesize
    \caption{Details of Computation and Sampling Cost Across Varying Search Space Sizes}
    \begin{tabular}{p{2.5cm}p{3.5cm}p{4.5cm}}
        \toprule
        \multicolumn{3}{c}{Active Discovery of Handwritten Digits} \\
        \midrule
        Search Space & Computation Cost & Sampling Time per observation step (Seconds)  \\
        \midrule
        28 $\times$ 28 & 0.78 GB & 0.83    \\
        128 $\times$ 128 & 1.51 GB & 1.87  \\
        \bottomrule
    \end{tabular}
    \label{tab: COMP}
\end{table}

\section{Code Link}
Our code and models are publicly available at this \href{https://github.com/KevinG396/EM_PTDM}{\textcolor{blue}{link}}.